%% file: arxiv.tex
\documentclass[10pt,twocolumn,letterpaper]{article}
\usepackage[dvipsnames]{xcolor}

\usepackage{iccv}
\usepackage{times}
\usepackage{epsfig}
\usepackage{graphicx}
\usepackage{amsmath}
\usepackage{amssymb}
\usepackage{subcaption}
\usepackage{tikz}
\usepackage{pgfplots}
\usepackage{pgfplotstable}
\usepackage[accsupp]{axessibility}
\usepackage{bm}
\usepackage{booktabs}
\usepackage{amsfonts}       
\usepackage{array}
\usepackage{multirow}
\let\checkmark\undefined % dingbat redefines it
\usepackage{dingbat}
\usepackage{xspace}

\usepackage[pagebackref=true,breaklinks=true,letterpaper=true,colorlinks,bookmarks=false]{hyperref}

\iccvfinalcopy

\newcommand{\para}[1]{\noindent{\bf{#1}}}
\newcommand{\PAR}[1]{\noindent{\bf{#1.}}}
\newcommand{\vpara}[1]{\vspace{1mm} \noindent{\bf{#1}}}

\newcommand{\ul}[1]{\underline{#1}}
\newcommand{\croconew}{CroCo~v2\xspace}
\newcommand{\ours}{CroCo-Stereo\xspace}
\newcommand{\oursflow}{CroCo-Flow\xspace}

\begin{document}

%%%%%%%%% TITLE
\title{\croconew: Improved Cross-view Completion Pre-training \\ 
for Stereo Matching and Optical Flow \vspace{-0.1cm}}

\author{Philippe Weinzaepfel ~~~~~~~~~~~~ Thomas Lucas ~~~~~~~~~~~~ Vincent Leroy ~~~~~~~~~~~~ 
\\[0.04cm] Yohann Cabon ~~~~~~~~~~~~ Vaibhav Arora ~~~~~~~~~~~~  Romain Br\'egier ~~~~~~~~~~~~ Gabriela Csurka \\[0.04cm] Leonid Antsfeld ~~~~~~~~~~~~ Boris Chidlovskii ~~~~~~~~~~~~ J\'er\^ome Revaud \\[0.1cm]
NAVER LABS Europe \\[-0.05cm]
{\small \url{https://github.com/naver/croco} }
\vspace{-0.4cm}
}

\maketitle
% Remove page # from the first page of camera-ready.
%\ificcvfinal\thispagestyle{empty}\fi

%%%%%%%%% ABSTRACT
\begin{abstract}
Despite impressive performance for high-level downstream tasks, self-supervised pre-training methods have not yet fully delivered on dense geometric vision tasks such as stereo matching or optical flow.
The application of self-supervised 
concepts, such as instance discrimination or masked image modeling, to geometric tasks is an active area of research.
In this work, we build on the recent cross-view completion framework,
a variation of masked image modeling that leverages a second view from the same scene which makes it well suited for binocular downstream tasks.
The applicability of this concept has so far been limited in at least two ways: (a) by the difficulty of collecting real-world image pairs -- in practice only synthetic data have been used -- 
and (b) by the lack of generalization of vanilla transformers to dense downstream tasks 
for which relative position is more meaningful than absolute position.
We explore three avenues of improvement. 
First, we introduce a method to collect suitable real-world image pairs at large scale.
Second, we experiment with relative positional embeddings and show that they enable vision transformers 
to perform substantially better.
 Third, we scale up vision transformer based cross-completion architectures, which is made possible by the use of large amounts of 
 data.
With these improvements, we show for the first time that state-of-the-art results on 
stereo matching and optical flow can be reached without using any classical task-specific techniques like 
correlation volume, iterative estimation, image warping or multi-scale reasoning, thus paving the way towards universal vision models.
\end{abstract}

%%%%%%%%% Introduction
\section{Introduction}
\label{sec:intro}

\begin{figure}
\includegraphics[width=\linewidth]{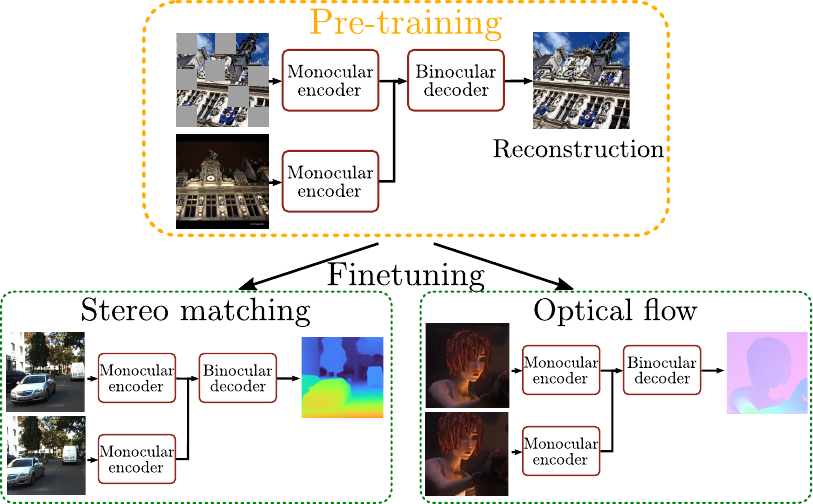} \\[-0.45cm]
\caption{\textbf{Pre-training for dense geometric tasks.} We pre-train a generic architecture, with a monocular encoder and a binocular decoder, with cross-view completion before finetuning it on the stereo matching or optical flow downstream task.}
\label{fig:teaser}

\vspace{-0.4cm}

\end{figure}

Self-supervised pre-training methods aim at learning rich representations from large amounts of unannotated data, which can then be finetuned on a variety of downstream tasks.
This requires the design of pretext tasks, for which supervision signal can be extracted from the data itself, as well as generic architectures that can be easily transferred.  
We hypothesize that successfully pre-training large models for geometric tasks such as stereo matching or optical flow, see Figure~\ref{fig:teaser}, requires three things all together: 
(a)~a well-designed dense pretext task inciting the understanding of 3D scene layout and geometry,  
(b)~an architecture that processes pairs of images, suitable for different downstream tasks,
and (c)~large-scale real-world data.
 
Early self-supervised methods proceeded by discarding part of the signal (\eg image color~\cite{ZhangECCV16ColorfulImageColorization}, patch ordering~\cite{jigsaw} or image orientation~\cite{rot}) and trying to recover it. 
Later methods based on instance discrimination~\cite{swav,dino,simclr,moco} were first to surpass supervised pre-training on high-level tasks: they are based on the idea that output features should be invariant to well-designed classes of augmentations. 
Another recently successful pretext task is masked image modeling (MIM)~\cite{msn,splitmask,mae,maskfeat,SimMIM,iBOT}, where part of the input data is masked and an auto-encoder is trained to restore the full signal from the remaining visible parts.
Instance
discrimination and MIM methods have achieved excellent performance on semantic tasks such as image classification, in particular with limited amounts of annotated data~\cite{msn,simclr2,fixmatch}, but have not led to breakthroughs in more geometric tasks like stereo matching and optical flow.

\begin{figure}
\includegraphics[width=\linewidth]{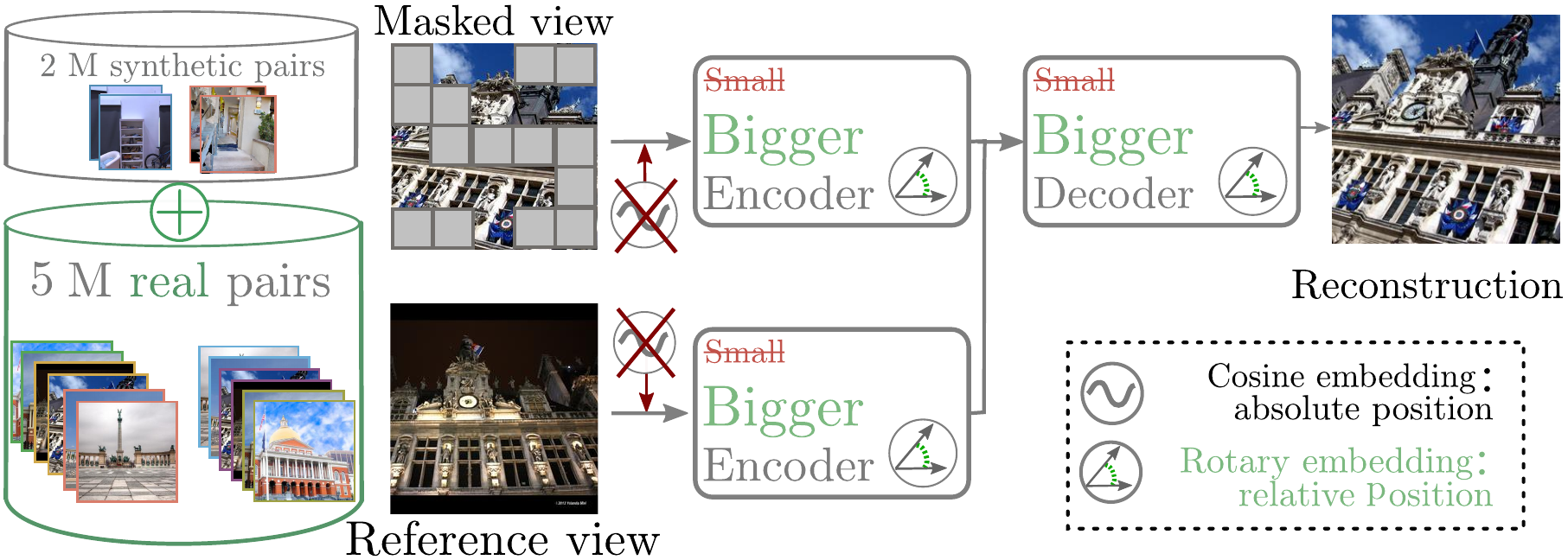} \\[-0.45cm]
\caption{\textbf{Overview of the improvements in \croconew for cross-view completion pre-training:} (a) collecting and using real-world images, (b) using rotary positional embeddings which model \textit{relative} token positions, instead of absolute positions using the standard cosine embedding, (c) increasing network size both in the encoder and the decoder.}
\label{fig:intro}
\vspace{-0.15cm}

\end{figure}

Adapting self-supervised pre-training to geometric vision tasks is an active area of research.
Attempts have been made to design contrastive learning objectives at the pixel or patch level~\cite{densecl,pixpro,loco}, but their performance gains have so far been more moderate than for global tasks.
Besides, these gains are mainly demonstrated for dense \emph{semantic} tasks such as semantic segmentation or object detection, rather than for geometric tasks such as depth estimation or stereo matching.
Recently,~\cite{croco} proposed the pretext task of cross-view completion (CroCo), a variant of MIM where a partially masked input image is reconstructed given visible patches and an additional view of the same scene.
This pre-training objective is well suited to geometric downstream tasks as (a) it leverages pairs of images and (b) extracting relevant information from the second view requires geometric understanding of the scene.
The CroCo architecture consists of a vision transformer (ViT)~\cite{vit} encoder to extract  features for the non-masked tokens of the first image, as well as for the second reference image, and a transformer to decode the features and reconstruct the masked image, as illustrated in Figure~\ref{fig:intro}. 

In spite of these advances, leveraging cross-view completion for geometric vision tasks remains challenging for at least two reasons.
First, training with cross-view completion requires image pairs depicting the same scene; this can be hard to acquire at scale, yet scale is the cornerstone of the success of self-supervised pre-training. 
In practice, the CroCo model of~\cite{croco} is pre-trained solely with synthetic data, which may limit its final performance.
Second, most models trained with masking rely on ViTs~\cite{vit}, which typically use absolute positional embeddings. 
These do not generalize well to new image resolutions when finetuning, and are not always robust to cropping. 
This limits the applicability of current cross-view completion methods and may explain why the downstream tasks presented in \cite{croco} mostly use low-resolution squared images.

\begin{figure*}
\centering
\newcommand{\figpairwidth}{0.115\linewidth}

\resizebox{\linewidth}{!}{
\begin{tabular}{@{}l@{~ }p{\linewidth}@{}}
\rotatebox{90}{\small Habitat (synth.)} & 
\includegraphics[width=\figpairwidth]{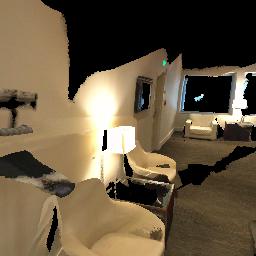}
\includegraphics[width=\figpairwidth]{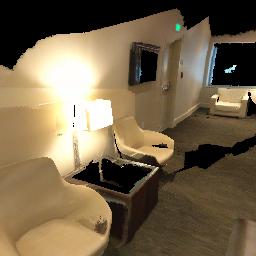} \hfill
\includegraphics[width=\figpairwidth]{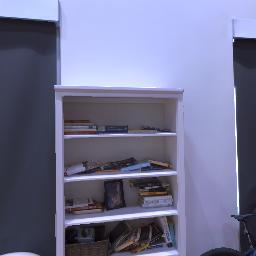}
\includegraphics[width=\figpairwidth]{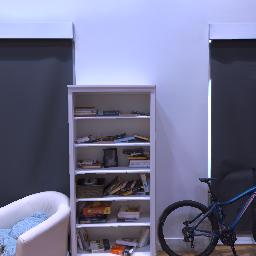} \hfill
\includegraphics[width=\figpairwidth]{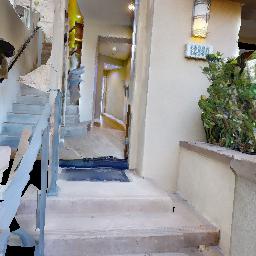}
\includegraphics[width=\figpairwidth]{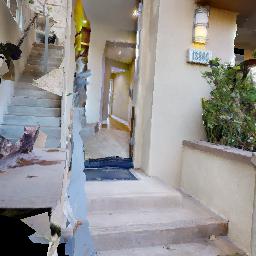} \hfill
\includegraphics[width=\figpairwidth]{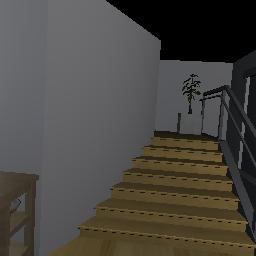}
\includegraphics[width=\figpairwidth]{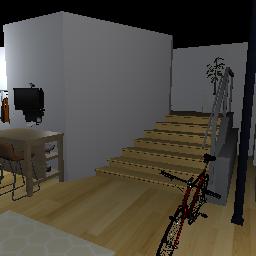} \\

\rotatebox{90}{\small ~~ARKitScenes} & 
\includegraphics[angle=-90,origin=c,width=\figpairwidth]{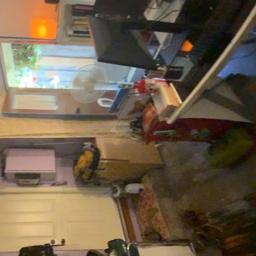}
\includegraphics[angle=-90,origin=c,width=\figpairwidth]{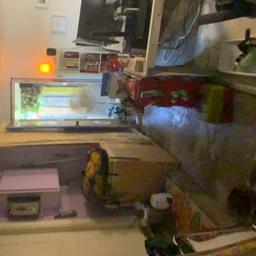} \hfill
\includegraphics[angle=0,origin=c,width=\figpairwidth]{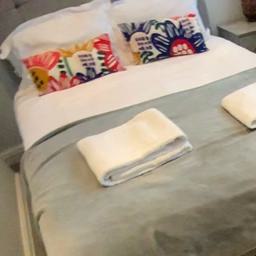}
\includegraphics[angle=0,origin=c,width=\figpairwidth]{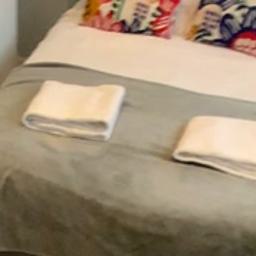} \hfill
\includegraphics[angle=0,origin=c,width=\figpairwidth]{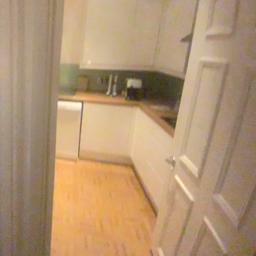}
\includegraphics[angle=0,origin=c,width=\figpairwidth]{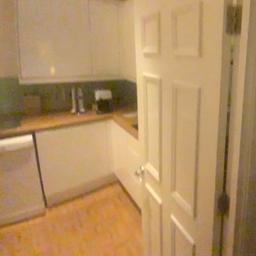} \hfill
\includegraphics[angle=-90,origin=c,width=\figpairwidth]{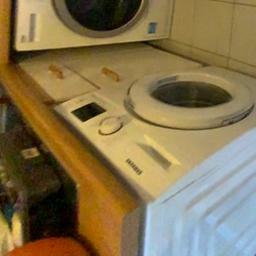}
\includegraphics[angle=-90,origin=c,width=\figpairwidth]{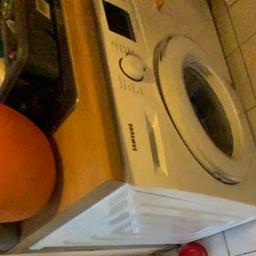} \\[-0.1cm]

\rotatebox{90}{\small ~~~MegaDepth} & 
\includegraphics[width=\figpairwidth]{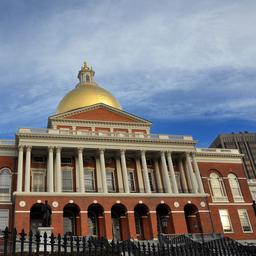}
\includegraphics[width=\figpairwidth]{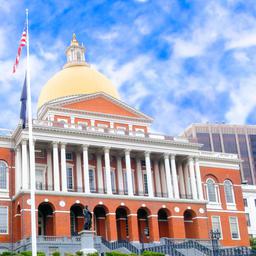} \hfill
\includegraphics[width=\figpairwidth]{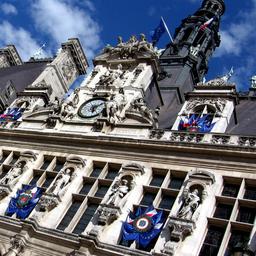}
\includegraphics[width=\figpairwidth]{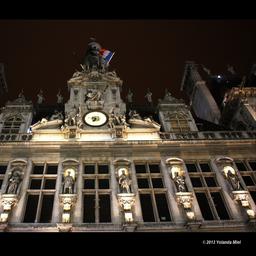} \hfill
\includegraphics[width=\figpairwidth]{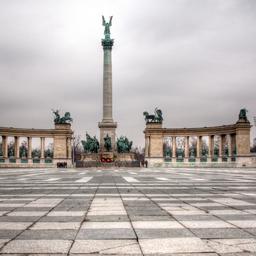}
\includegraphics[width=\figpairwidth]{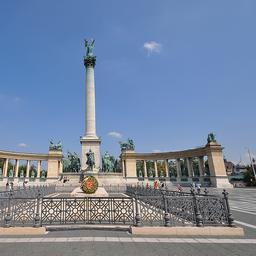}
\hfill
\includegraphics[width=\figpairwidth]{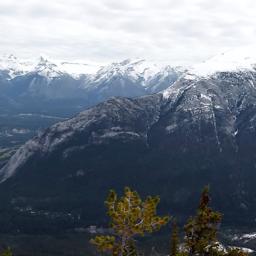}
\includegraphics[width=\figpairwidth]{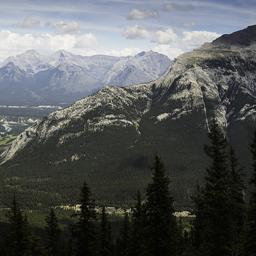}
\\[-0.1cm]

\rotatebox{90}{\small ~3DStreetView} &
\includegraphics[width=\figpairwidth]{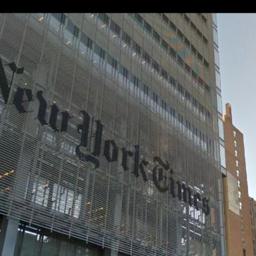}
\includegraphics[width=\figpairwidth]{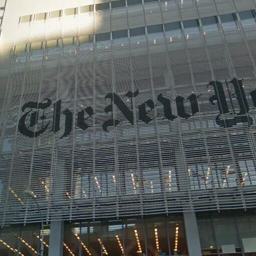} \hfill
\includegraphics[width=\figpairwidth]{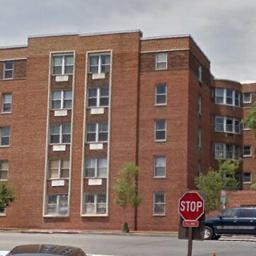}
\includegraphics[width=\figpairwidth]{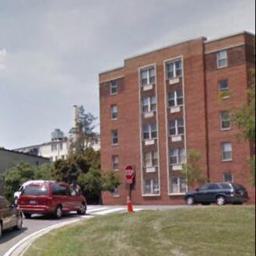} \hfill
\includegraphics[width=\figpairwidth]{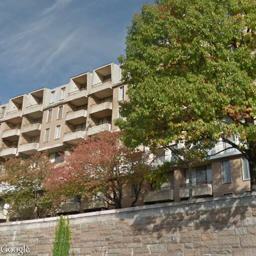}
\includegraphics[width=\figpairwidth]{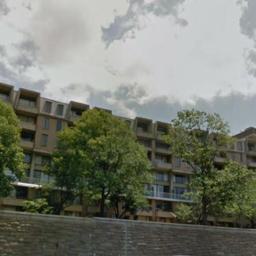} 
\hfill
\includegraphics[width=\figpairwidth]{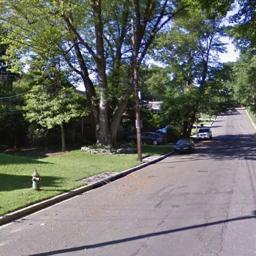}
\includegraphics[width=\figpairwidth]{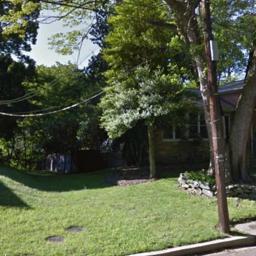} \\[-0.1cm]

\rotatebox{90}{\small ~~~~IndoorVL} & 
\includegraphics[width=\figpairwidth]{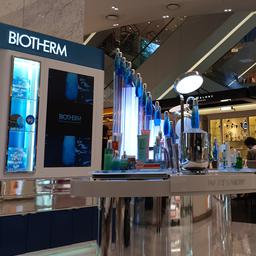}
\includegraphics[width=\figpairwidth]{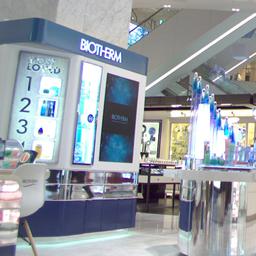} \hfill
\includegraphics[width=\figpairwidth]{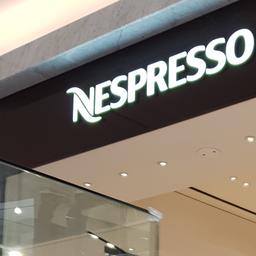}
\includegraphics[width=\figpairwidth]{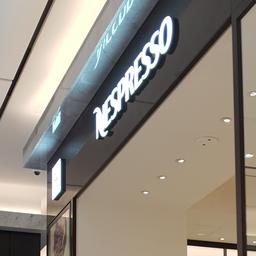} \hfill
\includegraphics[width=\figpairwidth]{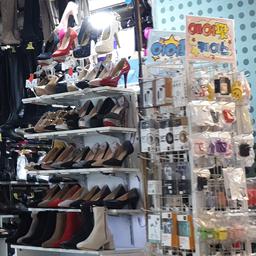}
\includegraphics[width=\figpairwidth]{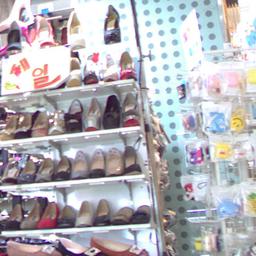} \hfill
\includegraphics[width=\figpairwidth]{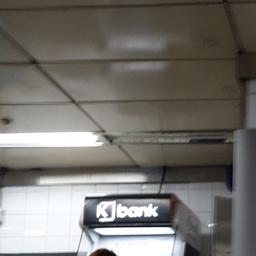}
\includegraphics[width=\figpairwidth]{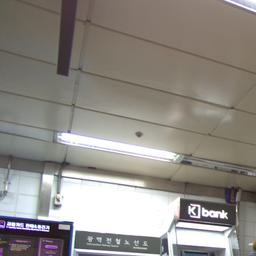} 
\\[-0.05cm]
\end{tabular}
}
\caption{\textbf{Example of pre-training cropped image pairs} from Habitat which was the synthetic data used by CroCo~\cite{croco} on the top row, and from real-world datasets we use in this paper (from ARKitScenes, MegaDepth, 3DStreetView and IndoorVL) below.}
\label{fig:pairs}

\vspace{-0.3cm}

\end{figure*}

In this paper, we propose solutions to these limitations that enable
to pre-train a large-scale cross-view completion model, see Figure~\ref{fig:intro}, leading to state-of-the-art performance on stereo matching and optical flow.
First, we tackle the problem of scalable pair collection, and gather millions of training pairs from different real-world datasets which cover various scenarios like indoor environments, street view data and landmarks, see Figure~\ref{fig:pairs}.
To generate high-quality pre-training pairs, we carefully control the visual overlap for each pair of images. 
In fact, pairs with high overlap make the task trivial, whereas pairs with negligible overlap reduce it to standard MIM~\cite{croco}.
To measure this overlap, we leverage extra information available such as 3D meshes, additional sensors like LIDAR, or Structure-from-Motion (SfM) reconstructions for datasets with sufficient image coverage.
From these data, 
we generate a set of high quality image pairs with sufficient overlap and viewpoint difference while also ensuring high diversity between pairs. 
Second, these large-scale datasets of pre-training pairs allow to scale up the model: (a) we use a larger encoder to extract better image features and (b) also scale up the decoder, which is responsible for combining information coming from the two views.
Third, instead of the standard cosine positional embedding which encodes absolute positional information, we rely on the Rotary Positional Embedding (RoPE)~\cite{rope} 
which efficiently injects relative positional information of token pairs in the attention mechanism. 

We finetune our pre-trained model, referred to as \croconew, with this improved cross-view completion scheme on stereo matching and optical flow using a Dense Prediction Transformer (DPT)~\cite{dpt} head. 
Our models, termed \ours and \oursflow, are simple and generic: 
we rely on a plain ViT encoder, followed by a plain transformer decoder which 
directly predicts the output (disparity for stereo, or optical flow) through the DPT head.
We believe this is a meaningful step towards a universal vision model, \ie, that can solve numerous vision tasks with a common architecture. In contrast to state-of-the-art methods for stereo matching or optical flow, our architecture does not rely on task-specific designs such as cost volumes~\cite{flowformer,jie2018left,gcnet,stereonet,yin2019hierarchical}, image warping~\cite{brox2004high,pwcnetplus}, iterative refinement~\cite{crestereo,raftstereo,raft} or multi-level feature pyramids~\cite{leastereo,crestereo,pwcnetplus}. While task-specific structures and prior knowledge may yield more data-efficient approaches, they come at the cost of being tailored to a single task. Our proposed pre-training allows us to eschew these and still reaches state-of-the-art performance on various stereo matching and optical flow benchmarks such as KITTI 2015~\cite{kitti15}, ETH3D~\cite{eth3d}, Spring~\cite{spring} or MPI-Sintel~\cite{sintel}.

%%%%%%%%% Related
\section{Related work}
\label{sec:related}

\PAR{Self-supervised learning}
The success of instance discrimination~\cite{swav,dino,simclr,byol,moco} has drawn a lot of attention to self-supervised learning in computer vision~\cite{JingPAMI21SelfSupervisedVisualFeatureLearningDeepSurvey}.
In that paradigm, variants of an image are obtained by applying different data augmentations. Features extracted from the different variants are trained to be similar, while being pushed away from features obtained from other images.
Such self-supervised models are particularly well tailored to image-level tasks, such as image classification, and have led to state-of-the-art performance on various benchmarks.
Recent studies suggest that this success could be due to the object-centric~\cite{purushwalkam2020demystifying} and the balanced~\cite{assran2022hidden} nature of ImageNet~\cite{imagenet} that is used for pre-training.
Recently, inspired by BERT~\cite{BERT} in natural language processing, different masked modeling methods have been adapted to computer vision. 
MIM pre-training aims at reconstructing masked information from an input image either in the pixel space~\cite{sit,multimae,iGPT,splitmask,mae,SimMIM}, or in the feature space~\cite{msn,data2vec,maskfeat}, and sometimes after quantization~\cite{BEiT,iBOT}.
Recent works combine this framework in a teacher-student approach~\cite{lee2022exploring,MST} with improved masking strategy~\cite{CiM,AttMask,MST}.
Overall, MIM models perform well on classification tasks. They have obtained some success on denser tasks such as object detection~\cite{mae} or human pose estimation~\cite{vitpose}, and have been applied to robotic vision~\cite{maerobot} when pre-trained on related datasets.
More recently, CroCo~\cite{croco} introduces the pretext task of cross-view completion, where a second view of the same scene is added to MIM. This is well suited to geometric downstream tasks: to leverage the second view and improve reconstruction accuracy, the model has to implicitly 
be aware of the geometry of the scene. CroCo outperforms MIM pre-training on an array of geometric tasks. However, it relies on synthetic data only, which may be sub-optimal, and does not reach the performance of the best task-specific methods.

\PAR{Positional embeddings} 
Since a ViT treats its input as an orderless set of image patches or tokens, positional embeddings are a necessary tool to keep track of the position of each patch token from the original image.
They can be either learned~\cite{dino,vit} or handcrafted, such as the cosine positional embeddings from the original transformer~\cite{attn}.
Both learned and cosine embeddings are added explicitly to the signal and contain absolute positional information.
However, models for pixel-level dense computer vision tasks should be able to process various image resolutions and be robust to cropping. Thus, relative positional embeddings, \eg~\cite{ShawNAACL18RelPos}, that consider distances between tokens are preferable. 
For instance, Bello~\etal~\cite{BelloICCV19AttentionAugmentedCNN} 
achieve better object detection results
using relative self-attention.
Similarly, 
Swin Transformers~\cite{swin} and Swin V2~\cite{swinv2} observed improved performance using relative positional embeddings, while~\cite{craft} showed it to be crucial in the cross attention for optical flow. 
Recently, \cite{rope} introduced the Rotary Positional Embedding (RoPE): 
a transformation to each key and query features is applied according to their absolute position, in such a way that the pairwise similarity scores used in the attention computation only depend on the relative positions of the token pairs and on their feature similarity. 
RoPE thus models relative positions 
at any resolution.

\para{Stereo matching and optical flow} can both be seen as a dense correspondence matching problem~\cite{unimatch}. However the priors about matching itself and the completion of unmatched regions differ. This explains why most models are dedicated to one specific task despite many similarities in the strategies~\cite{laga2020survey,zhaisurvey21}.
Dense matching is most often posed with correlation/cost volume estimation from which matches can be extracted~\cite{flownet,sceneflow}.
For stereo, this volume typically has three dimensions~\cite{jie2018left,stereonet,yin2019hierarchical}, the third dimension representing a discretization of the disparity level, or four dimensions~\cite{psmnet,gcnet,nie2019multi}. 
For optical flow, each pixel of the first image can be associated to any pixel of the second, resulting in a 4D correlation volume. The complexity of building, storing and leveraging such volume motivated numerous methods revolving around the ideas of coarse-to-fine~\cite{deqflow,raft,gmflownet}, warping~\cite{pwcnetplus}, sparse formulation~\cite{jiang21}, random search~\cite{dip}, dimension separation~\cite{sepflow}, tokenization~\cite{flowformer}. 
Interestingly, recent  works~\cite{craft,gmflow,unimatch} leverage cross-attention to facilitate inter-image information exchanges but still rely on a low-resolution correlation volume, followed by an iterative refinement similar to~\cite{raft}. 
Unimatch~\cite{unimatch} made an important step towards a unified architecture for flow and stereo, but still relies on task-dependent (a) cross-attention mechanisms, (b) correlation volume and (c) post-processing. 
We similarly use the same architecture for both tasks, but our standard transformer model without cost volume can be pre-trained with existing self-supervised approaches and directly finetuned as is.

Several works propose self-supervised methods for estimating depth using stereo pairs or videos~\cite{godard2017unsupervised,godard2019digging}, stereo with matching priors~\cite{zhou2017unsupervised}, or optical flow~\cite{sun2014quantitative,selflow,yu2016back} typically with an unsupervised reconstruction loss. The main difference between this paradigm and ours is that we aim to pre-train a task-agnostic model that can be finetuned to different tasks, while these approaches aim to remove supervision for a single task.

%%%%%%%%% Method
\section{Cross-view completion pre-training at scale}
\label{sec:method}

Our proposed pre-training method is based on the recently introduced cross-view completion (CroCo) framework~\cite{croco}.
It extends MIM 
to pairs of images.
Given two different images depicting a given scene, the two images are divided into sets of non-overlapping patches, denoted as tokens, and 90\% of the tokens from the first image are masked.
The remaining ones are fed to a ViT~\cite{vit} encoder to extract features for the first image. Similarly, tokens from the second image are fed to the same encoder with shared weights,
and a ViT decoder processes the two sets of features together to reconstruct the target.
Figure~\ref{fig:intro} provides an overview of the pre-training stage.
Compared to standard masked image modeling methods, this approach can leverage the information in the second view to resolve some of the ambiguities about the masked context.
To leverage this information, the model has to implicitly reason about the scene geometry and the spatial relationship between the two views, which primes it well for geometric tasks. 

\begin{figure*}[t]
\centering
\includegraphics[width=\linewidth]{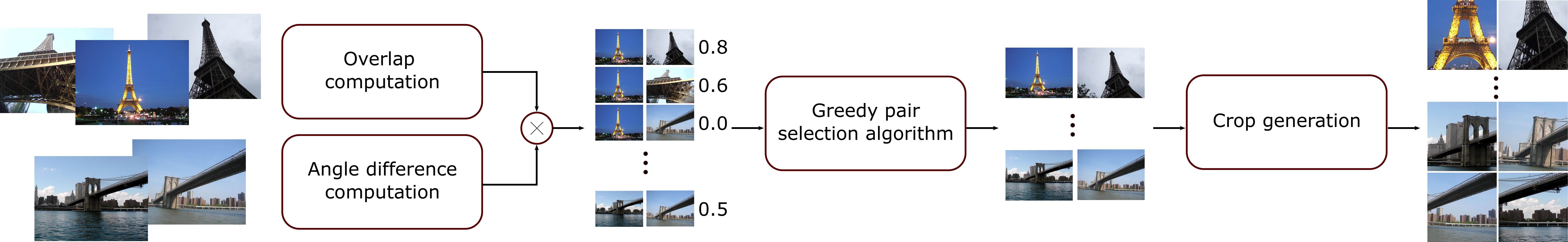} \\[-0.1cm]
\caption{\textbf{Overview of our pre-training cropped image pair collection method.} Given a dataset of posed images, optionally with point clouds (\eg from SfM) or meshes of the scene, we first measure the visual overlap between pairs and the viewpoint angle difference. Based on these scores, we use a greedy algorithm to select diverse image pairs and finally generate crops from them.}

\vspace{-0.4cm}

\label{fig:pipeline}

\end{figure*}

\PAR{Training data} 
Collecting pairs of images that are suitable for this approach is non-trivial.
First, images have to be paired together without manual annotation; second, their visual overlap has to be carefully controlled for the pairs to be useful. 
In~\cite{croco}, only synthetic data generated with the Habitat simulator~\cite{habitat} is used, which restricts the variety of the pre-training data. 
In contrast, we propose an approach to this real-world image pairing problem, necessary to use cross-view completion at scale, as detailed in  Section~\ref{sub:pairs}.

\PAR{Positional embeddings} The architecture used in~\cite{croco} adapts ViTs to process pairs of images, 
by using cross-attention inside the decoder.
Following standard practices, in their work cosine positional embedding is added to the token features prior to the encoder and the decoder. This models absolute position while dense tasks must typically be robust to cropping or images of various resolutions.
In Section~\ref{sub:posembed}, we describe how relative positional embeddings can be adapted to cross-view completion.

\PAR{Large-scale models} Finally, we discuss scaling-up the model in Section~\ref{sub:decoder}. CroCo~\cite{croco} uses a ViT-Base encoder (12 blocks, 768-dimensional features, 12 attention heads) and a decoder composed of 8 blocks with 512-dimensional features and 16 heads.
Using our large-scale dataset of real-world image pairs, we are able to scale to larger ViT architectures and demonstrate consistent performance gain. 

\subsection{Collecting real-world image pairs}
\label{sub:pairs}

We now present our approach to automatically select image pairs
from real-world datasets that are suitable for pre-training. 
To be useful, pairs need to depict the same scene with some partial overlap.
The overlap should not be small to the point where the task boils down to auto-completion.
It should not be high either to the point where the task becomes a trivial `copy and paste', \ie, without requiring any understanding of scene geometry. 
On top of that, diversity should be as high as possible among pairs. 
We propose to use datasets that offer ways of getting information about the geometry of the scene and the camera poses.
This signal can be captured using additional sensors like LIDAR, or it can be extracted using structure-from-motion (SfM) techniques if the images offer enough coverage of the scene. 
We use this information to obtain an image pair quality score based on overlap and difference in viewpoint angle.
We then use a greedy algorithm to select a diverse set of image pairs. Finally, we generate overlapping image crops by leveraging image matching. Figure~\ref{fig:pipeline} gives an overview of our approach and we detail each step below.

\PAR{Computing overlap scores} 
The first step is to compute overlap scores for candidate pairs. 
We develop several approaches depending on the available information.

$\circ \hspace{0.1cm} \textit{ARKitScenes}$
\cite{arkit} provides 450,000 frames from 1,667 different indoor environments. The availability of the corresponding mesh for each frame enables the computation of the overlap between every pair of images.
For each image $I$, we retrieve the set of mesh vertices $\mathcal{P}(I)$ that are visible. 
We then measure the intersection-over-union (IoU) of the vertices (3D points) for each pair of images $(I_1,I_2)$ as:
\begin{equation}
 IoU(I_1,I_2) = \frac{\vert \mathcal{P}(I_1) \cap \mathcal{P}(I_2) \vert}{\vert \mathcal{P}(I_1) \cup \mathcal{P}(I_2) \vert}.
\end{equation}

$\circ \hspace{0.1cm} \textit{MegaDepth}$ 
\cite{megadepth} consists of around 300,000 images downloaded from the web corresponding to 200 different landmarks.
From these images, a point cloud model for each landmark obtained using structure-form-motion (SfM) with COLMAP~\cite{colmapsfm} is also provided.
As above, it is possible to measure the vertex-based IoU between pairs of images, where each vertex is in this case a 3D point from the point cloud.
Unfortunately, occlusions cannot be taken into account due to the absence of 3D mesh, which greatly degrades the overlap estimation.
We propose a simple yet effective solution: we create an artificial occlusion model by attaching a ball of fixed radius to each 3D point, which occludes the vertices placed behind it.
This way, we can compute a set of visible vertices for each image and evaluate the IoU as done previously. 

$\circ \hspace{0.1cm} \textit{3D Street View}$ 
\cite{streetview} contains 25 million street view images from 8 cities. 
In addition to the camera pose, the 3D location and orientation (normal vector) of the target buildings are provided.
To compute the overlap score, we create a pseudo 3D point cloud and apply the same technique as for MegaDepth.
We start from an empty point cloud and append, for each target building, a $10 \times 6$ meters grid of $7 \times 11$ balls oriented according to the provided annotation.

$\circ \hspace{0.1cm} \textit{Indoor Visual Localization datasets}$ 
(IndoorVL)~\cite{gangnam} contains over 135,000 images from a large shopping mall and a large metro station in Seoul, South Korea, 
captured regularly with several months interval 
with 
10 cameras and 2 laser scanners. 
The data is provided with accurate camera poses obtained via LiDAR SLAM refined by SfM-based optimization.
We directly measure the overlap between images using the intersection between the camera frustrums using the accurate camera poses provided with the dataset.
To encourage further diversity, we multiply this 
score by a factor $0.8$ if both images come from the same capture session, thus favoring pairs taken with several months interval.

\PAR{Greedy image pair selection}
We rely on the overlap scores described above to select high quality pairs. 
This is however not sufficient: we also need pairs to be diverse, which would not be the case when randomly selecting good pairs, as images in the dataset can be very correlated. Therefore, we use a greedy algorithm to select non-redundant image pairs for pre-training. 
First, for each image pair $(I_1,I_2)$ we use a quality pair score $s$ given by:
\begin{equation}
    s(I_1,I_2) = IoU(I_1,I_2) \times 4 \cos(\alpha) \big(1-\cos(\alpha)\big),
\end{equation}
where $\alpha$ denotes the viewpoint angle difference between the two images (all the datasets above provide camera poses).
The function $4 \cos(x) (1-\cos(x))$ has a maximum value of $1$ for $x=60^{\circ}$, $0$ value for $x=0^{\circ}$ and $x=90^{\circ}$, and it is negative for angles above $90^{\circ}$. This score thus favors pairs with different viewpoints while still having large overlaps. Given the score for every pair, we aim at building a large number of image pairs while ensuring diversity, \ie, avoiding content redundancy. 
To do this, we use a greedy algorithm, where each time we select a pair of images with maximum score, 
we discard the two images forming the pair, as well as images that have too large IoU (above $0.75$) with any of the two. We iteratively repeat this process until there is no pair with a score above a certain threshold. 

\PAR{Crop generation per pair}
For pre-training, we use fixed-size crops of $224{\times}224$ pixels, as considering 
higher-resolution images would be too costly. 
In practice, we generate $256{\times}256$ crops and apply random cropping during pre-training.
To generate crops on pairs of images while maintaining overlaps, we rely on quasi-dense keypoint matching, namely DeepMatching~\cite{deepmatching}, except for pairs from ARKitScenes where we directly use matches from the mesh.
Given the matches, we consider a grid of crops in the first image, estimate the corresponding matching crop in the second image and keep 
those with the most consistent matches and without overlap in the first image.

\begin{figure*}[t]
\includegraphics[width=\linewidth]{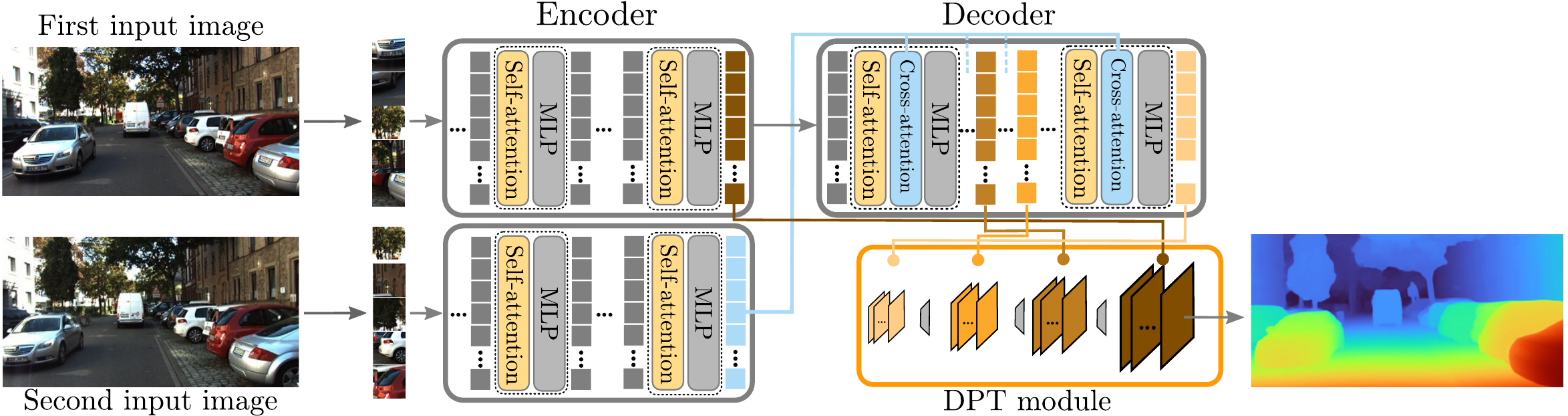} \\[-0.5cm]
\caption{\textbf{Architecture of \ours and \oursflow.} The two images (left and right views for stereo, two frames for flow) are split into patches and encoded with a series of transformer blocks with RoPE positional embeddings. The decoder consists in a series of transformer decoder blocks (self-attention among token features from the first image, cross-attention with the token features from the second image, and an MLP). Token features from different intermediate blocks are fed to the DPT module~\cite{dpt} to obtain the final prediction.}
\label{fig:stereo}

\vspace{-0.35cm}

\end{figure*}

\PAR{Overall statistics}
In total, we collected about 5.3 million real-world pairs of crops with the process described above, with respectively 1,070,414 pairs from ARKitScenes~\cite{arkit}, 2,014,789 pairs from MegaDepth~\cite{megadepth}, 655,464 from 3DStreetView~\cite{streetview}, and 1,593,689 pairs from IndoorVL~\cite{gangnam}. We added this to 1,821,391 synthetic pairs generated with the Habitat simulator~\cite{habitat}, following the approach of~\cite{croco}. 
Example pairs for each dataset are shown in Figure~\ref{fig:pairs}.
They cover various scenarios, from indoor rooms -- synthetic with Habitat or real with ARKitScenes -- to larger crowded indoor environment (IndoorVL), landmarks (MegaDepth) and outdoor streets (3DStreetView).

\subsection{Positional embeddings}
\label{sub:posembed}

We replace the cosine embeddings, which inject absolute positional information, by Rotary Positional Embedding (RoPE)~\cite{rope}.
RoPE efficiently injects information about the \emph{relative} positioning of feature pairs when computing attention.
Formally, let $q$ and $k$ represent a query and a key feature, at absolute positions $m$ and $n$ respectively.
The main idea of RoPE is to design an efficient function $f(x,p)$ that transforms a feature $x$ according to its absolute position $p$ such that the similarity between the transformed query and the transformed key $\langle f(q,m), f(k,n) \rangle$ is a function of $q$, $k$ and $m-n$ only.
\cite{rope} showed that a simple transformation such as applying rotations on pairs of dimensions according to a series of rotation matrices at different frequencies satisfy this desirable property.
To deal with 2D signals 
such as images, we 
split the features into 2 parts, we apply the 1D positional embedding of the x-dimension on the first part, and the embedding of the y-dimension on the second part.

\subsection{Scaling up the model}
\label{sub:decoder}

The combination of information 
extracted from the two images only occurs in the decoder.
Following MAE~\cite{mae}, CroCo~\cite{croco} uses a small decoder of 8 blocks consisting of self-attention, cross-attention and an MLP, with 512 dimensions and 16 attention heads. As the decoder is crucial for binocular tasks such as stereo or flow, we scale up the decoder and follow the ViT-Base hyper-parameters with 12 blocks, 768-dimensional features and 12 heads. We also scale up the image encoder from ViT-Base to ViT-Large, \ie, increase the depth from 12 to 24, the feature dimension from 768 to 1024 and the number of heads from 12 to 16. 

\PAR{Pre-training detailed setting}
We pre-train the network for 100 epochs with the AdamW optimizer~\cite{adamw}, a weight decay of $0.05$, a cosine learning rate schedule at a base learning rate of $3.10^{-4}$ with a linear warmup in the first 10 epochs, and a batch size of 512 spread on 8 GPUs.
During pre-training, we simply use random crops and color jittering as data augmentation.
We mask 90\% of the tokens from the first image.
Examples of cross-view completion obtained with our model are shown in Appendix~\ref{app:crocovis}.

\begin{table*}
\centering
\resizebox{\linewidth}{!}{
\begin{tabular}{ccccl@{\hskip 0.2cm}rrrrp{0cm}rrrr}
     \toprule
pos. & \multirow{2}{*}{encoder} & \multirow{2}{*}{decoder} & pre-train & & \multicolumn{4}{c}{Stereo (bad@1.0px$\downarrow$)} & & \multicolumn{4}{c}{Flow (EPE$\downarrow$)}  \\
emb. &  &  & data & & \multicolumn{1}{c}{\small Md} & \multicolumn{1}{c}{\small ETH} & \multicolumn{1}{c}{\small SF(c)} & \multicolumn{1}{c}{\small SF(f)} & & \multicolumn{1}{c}{\small FT(c)} & \multicolumn{1}{c}{\small FT(f)} & \multicolumn{1}{c}{\small Si.(c)} & \multicolumn{1}{c}{\small Si.(f)}  \\
\cmidrule(l){1-4} \cmidrule(lr){6-9} \cmidrule(lr){11-14}
cosine & ViT-B & Small & 2M habitat             & {\small (CroCo~\cite{croco})}  & 26.3 &  1.82 &   6.7 &   7.0 &       &3.89 &  3.56 &  2.07 &  2.57 \\
RoPE   & ViT-B & Small & 2M habitat             &                 & 25.3 &  \ul{0.60} &   6.0 &   6.3 &       &3.73 &  3.37 &  2.13 &  2.77 \\
RoPE   & ViT-B & Small & 2M habitat + 5.3M real &                 & 20.7 &  0.82 &   5.8 &   6.1 &       &3.35 &  2.94 &  1.76 &  2.30 \\
RoPE   & ViT-B & Base  & 2M habitat + 5.3M real &                 & \ul{17.1} &  1.14 &   \ul{5.3} &   \ul{5.6} &       &\ul{3.10} &  \ul{2.73} &  \ul{1.51} &  \bf{1.99} \\
  RoPE & ViT-L & Base  & 2M habitat + 5.3M real & {\small (\textbf{\croconew})} & \bf{15.5} &  \bf{0.38} &   \bf{5.0} &   \bf{5.3} &       &\bf{2.85} &  \bf{2.45} &  \bf{1.43} &  \bf{1.99}    \\
\bottomrule
\end{tabular}
} 
\caption{\textbf{Ablative study} of each change to CroCo with the percentage of pixels with error above 1px  (bad@1.0) on validation sets from Middlebury (Md), ETH3D, SceneFlow (SF) in clean (c) and final (f) renderings for stereo, and with the endpoint error (EPE) on validation sets from FlyingThings (FT) and MPI-Sintel (Si.) in both clean (c) and final (f) renderings for optical flow. A \textit{Small} decoder has 8 decoder blocks with 16 attention heads on 512-dimensional features, while the \textit{Base} one has 12 blocks with 12 heads on 768-dimensional features.
}
\label{tab:changes}
\end{table*}

\begin{table*}
\resizebox{\linewidth}{!}{
\begin{tabular}{lcccccccccccccccc}
\toprule
                        & {\small Bicyc2}    & {\small Compu}     & {\small Austr}     & {\small AustrP}    & {\small Djemb}     & {\small DjembL}    & {\small Livgrm}    & {\small Plants}    & {\small Hoops} & {\small Stairs} & {\small Nkuba} & {\small Class} & {\small ClassE} & {\small Crusa}  & {\small CrusaP} & \bf{avg}$\downarrow$ \\  
$\text{nd} < \text{400px}$ & \checkmark & \checkmark & \checkmark & \checkmark & \checkmark & \checkmark & \checkmark & \checkmark & & & & & \\
\midrule
LEAStereo~\cite{leastereo}    & 1.83 & 3.81 & 2.81 & 2.52 & 1.07 & 1.64 & 2.59 & 5.13 & 5.34 & 2.79 & 3.09 & 2.46 & 2.75 & 2.91 & 3.09 & 2.89 \\
AdaStereo~\cite{adastereo}    & 2.19 & 2.29 & 4.37 & 3.08 & 1.40 & 1.64 & 3.93 & 7.58 & 4.46 & 2.67 & 3.69 & 3.29 & 3.35 & 3.78 & 2.94 & 3.39 \\
HITNet~\cite{hitnet}          & 1.43 & 1.87 & 3.61 & 3.27 & 0.90 & 9.12 & 2.37 & 4.07 & 4.45 & 3.38 & 3.45 & 2.43 & 3.20 & 4.67 & 4.74 & 3.29 \\
RAFT-Stereo~\cite{raftstereo} & \ul{0.90} & 1.13 & 2.64 & \ul{2.22} & \bf{0.63} & 1.22 & 3.13 & 3.55 & 3.54 & \ul{1.89} & 4.36 & \bf{1.46} & 2.44 & 4.58 & 6.00 & 2.71 \\
CREStereo~\cite{crestereo}    & 1.38 & \bf{1.06} & 2.63 & 2.53 & \ul{0.64} & \bf{1.11} & \bf{1.42} & 5.31 & \ul{3.22} & 2.40 & 2.51 & \ul{1.92} & \ul{2.31} & \ul{1.78} & \ul{1.83} & \ul{2.10} \\
GMStereo~\cite{unimatch}      & 1.34 & \ul{1.32} & \ul{2.26} & 2.23 & 1.01 & 1.62 & \ul{1.84} & \ul{2.49} & \bf{3.19} & 2.18 & \ul{2.10} & 2.19 & \bf{2.08} & \bf{1.71} & \bf{1.75} & \bf{1.89} \\
\bf{\ours}             & \bf{0.84} & 1.45 & \bf{1.87} & \bf{1.83} & 0.69 & \ul{1.19} & 2.40 & \bf{2.28} & 8.31 & \bf{1.44} & \bf{1.96} & 3.99 & 4.61 & 2.48 & 2.81 & 2.36 \\
\bottomrule
\end{tabular}
}
\caption{\textbf{Evaluation on Middlebury} with the average error over all pixels for each sequence and the average (last column). Sequences are ordered according to their `nd' value, which is the official threshold of maximum disparity used to clip predictions before evaluation.}
\label{tab:middlebury2}

\vspace{-0.4cm}

\end{table*}

%%%%%%%%%% Stereo
\section{Application to stereo matching and flow}
\label{sec:stereo}

We now present \ours and \oursflow, our ViT-based correlation-free architecture for stereo matching and optical flow respectively, pre-trained with cross-view completion.
This is much in contrast to current state-of-the-art methods which rely on task-specific design in the form of cost volumes~\cite{flowformer,jie2018left,gcnet,stereonet,adastereo,pwcnetplus,acvnet,yin2019hierarchical,dip}, image warping~\cite{brox2004high,pwcnetplus}, iterative refinement~\cite{crestereo,raftstereo} and multi-level feature pyramids~\cite{leastereo,crestereo,pwcnetplus,hitnet}. Both \ours and \oursflow share the same architecture.

\PAR{Architecture} When finetuning the model for stereo or flow, both images are fed to the encoder as during pre-training (but without masking), and the decoder processes the tokens of both images.
To output a pixel-wise prediction, we rely on DPT~\cite{dpt}, which adapts the standard up-convolutions and fusions from multiple layers 
used in fully-convolutional approaches for dense tasks, to vision transformers.
This allows to combine features from different blocks by reshaping them to different resolutions and fusing them with convolutional layers. 
In practice, we use the features from $4$ blocks, regularly spread by an interval of a third of the decoder depth, starting from the last block, resulting in $1$ block at the end of the encoder and $3$ decoder blocks.

\PAR{Loss} 
We parameterize the output of the network with a Laplacian distribution~\cite{laplacian}: given an input pair $(\bm{x}_1, \bm{x}_2)$, the model outputs a location parameter $\mu_i$ and a scale parameter $d_i$ per pixel location $i$ and is trained to minimize the negative log-likelihood of the ground-truth target disparity, denoted $\bar{\mu}$, under the predicted distribution:
\begin{equation}
    - \log p(\bar{\mu}| \mu, d) =  \sum_i \left[ \frac{\vert \mu_i - \bar{\mu}_i \vert}{d_i} -  2 \log d_i \right]. 
    \label{eq:loss}
\end{equation}
The scale parameter $d$ can be interpreted as an uncertainty score for the prediction: large errors are penalized less when $d$ is high, while good predictions are rewarded more if $d$ is low. 
It is thus optimal for the network to adapt the scale parameter. 
The second term comes from the normalization term of the Laplacian density and avoids the degenerate solution of always predicting a low scale parameter.
Empirically, we find that using a probabilistic loss improves performance, see Appendix~\ref{appsub:loss} for the ablation, and is useful for tiling strategies during inference, because it provides a per-pixel confidence estimate, as detailed below.
A parameterization of $d_i$ ensures its positiveness: for stereo matching we use $d_i = e^{2 \alpha (\sigma(d'_i / \alpha)-0.5)}$, with $\sigma$ the sigmoid function and $\alpha=3$, and for optical flow $d_i = 1/\beta + (\beta-1/\beta) \sigma(d'_i)$ with $\beta=4$, unless otherwise stated.

\PAR{Training} We train \ours using $704{\times}352$ crops from various stereo datasets: CREStereo~\cite{crestereo}, SceneFlow~\cite{sceneflow}, ETH3D~\cite{eth3d}, Booster~\cite{booster}, Middlebury (2005, 2006, 2014, 2021 and v3)~\cite{middlebury}. We train \oursflow using $384{\times}320$ crops from the TartanAir~\cite{tartan}, MPI-Sintel~\cite{sintel}, FlyingThings~\cite{sceneflow} and FlyingChairs~\cite{flownet} datasets.
We refer to Appendix~\ref{app:details} for more details on these datasets, the splits we use for the ablations, the data augmentation strategy, as well as training hyper-parameters.

\PAR{Inference} We use a tiling-based approach.
We sample overlapping tiles with the same size as the training crops in the first image.
For each tile, we create a pair by sampling a corresponding tile at the same position from the second image.
We then predict the disparity or flow between each pair of tiles.
Such tiling approach was used \eg in~\cite{flowformer}. To merge the predictions done at a given pixel, we use a weighted average with weights $e^{- 2 \eta \alpha (\sigma(d'_i / \alpha) - 0.5)}$ with $\eta=5$ for stereo matching and $\alpha=5, \eta=2$ for optical flow, where $d'_i$ is the uncertainty predicted by the model.

%%%%%%%%%%%%%%%%% xp
\section{Experiments}
\label{sec:xp}

\PAR{Ablations}
We perform our ablations on the validation pairs (see Appendix~\ref{app:details} for the splits we use) of Middlebury, ETH3D and SceneFlow for stereo matching, and FlyingThings and MPI-Sintel for optical flow.
Table~\ref{tab:changes} reports the impact of the changes in \croconew to improve CroCo~\cite{croco} (pre-training data, positional embedding, larger encoder and decoder).
We observe that they all lead to consistent improvements: replacing the cosine absolute positional embedding by RoPE, scaling up the decoder, using larger-scale pre-training data and a larger encoder. Altogether, this allows \eg to improve performance as measured by the bad@1.0px metric from $26.3$ to $15.5$ on Middlebury (stereo matching), or the EPE from $2.07$ to $1.43$ on MPI-Sintel in its clean rendering (optical flow).

To further benchmark \croconew, we evaluate the pre-training of the encoder only on monocular tasks following the protocol of~\cite{multimae}. For semantic segmentation on ADE20k~\cite{ade}, we obtain 44.7 mean Intersection over Union \vs 40.6 for CroCo~\cite{croco}, and for monocular depth estimation on NYU v2~\cite{nyuv2}, we obtain 93.2 delta-1 \vs 90.1 for~\cite{croco}.

We provide in Appendix~\ref{app:morexp} an ablation on the impact of pre-training (\ie, a comparison with a randomly initialized network for finetuning), an ablation on the masking ratio during pre-training as well as a comparison between the L1 loss and Laplacian loss during finetuning.

\PAR{\ours \vs the state of the art} 
We now evaluate \ours on the official leaderboards of Middlebury~\cite{middlebury}, KITTI 2015~\cite{kitti15}, ETH3D~\cite{eth3d} and Spring~\cite{spring}.

\begin{figure}
\newcommand{\exlength}{0.245\linewidth}
\resizebox{\linewidth}{!}{
\begin{tabular}{c@{ }c@{ }c@{ }c@{ }c}
{\small Left image} & {\small Ground truth} & 
{\small CREStereo~\cite{crestereo}} & {\small \bf{\ours}} \\
\includegraphics[width=\exlength]{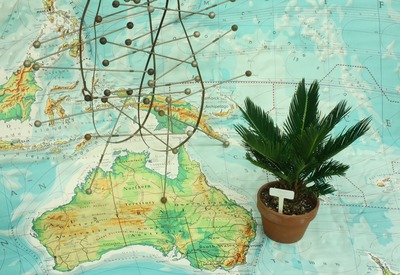} &
\includegraphics[width=\exlength]{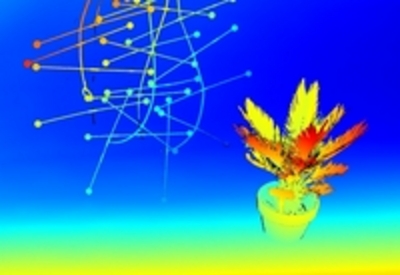} &
\includegraphics[width=\exlength]{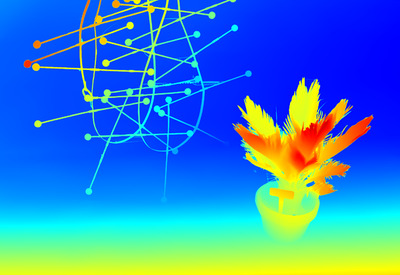} & 
\includegraphics[width=\exlength]{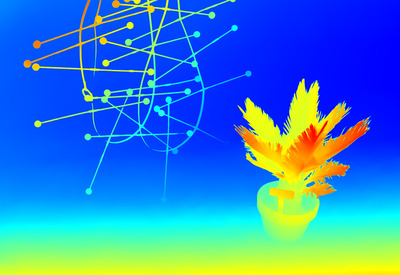} \\[-0.1cm]
\includegraphics[width=\exlength]{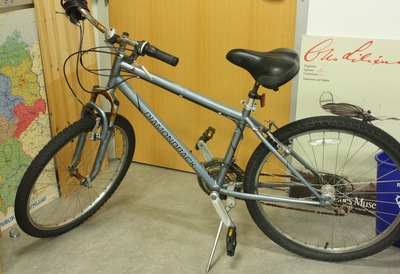} & 
\includegraphics[width=\exlength]{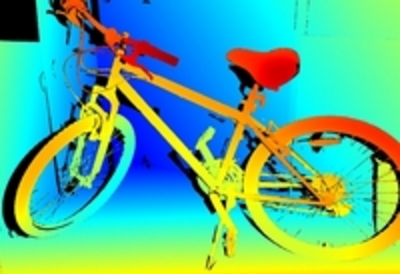} & 
\includegraphics[width=\exlength]{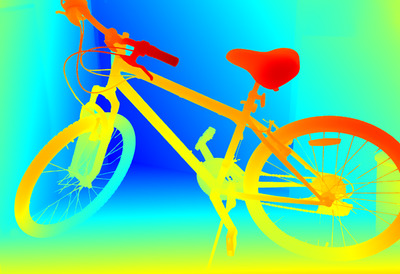} &
\includegraphics[width=\exlength]{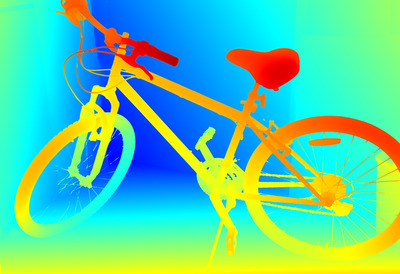} \\[-0.1cm]
\includegraphics[width=\exlength]{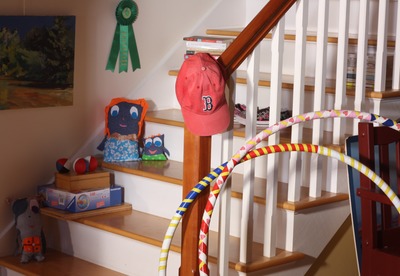} &
\includegraphics[width=\exlength]{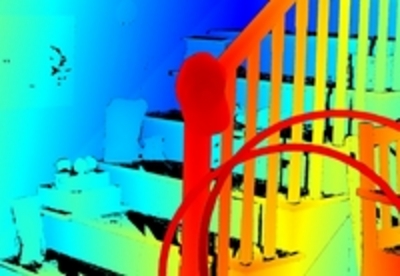} &
\includegraphics[width=\exlength]{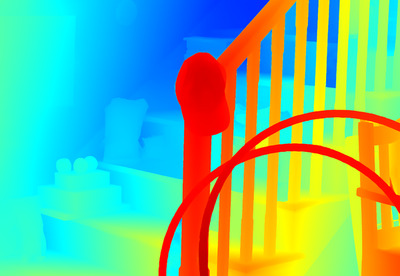} &
\includegraphics[width=\exlength]{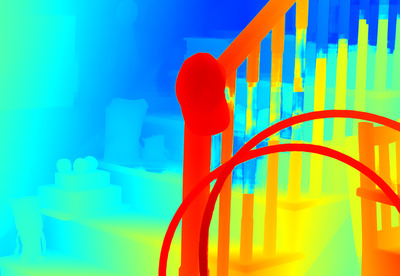} 
\end{tabular}
} \\[-0.25cm]
\caption{\textbf{Three example results from the Middlebury test set} (Australia, Bicycle2 and Hoops) with from left to right: the left image, the ground truth, 
CREStereo~\cite{crestereo} and \ours.}
\label{fig:examples}
\end{figure}

\begin{table}[]
 \centering
 \small
 %\resizebox{\linewidth}{!}{
 \begin{tabular}{lccc}
 \toprule
 Method & D1-bg$\downarrow$ & D1-fg$\downarrow$ & D1-all$\downarrow$ \\
 \midrule
 AdaStereo~\cite{adastereo} & 2.59      & 5.55      & 3.08 \\
 HITNet~\cite{hitnet}       & 1.74      & 3.20      & 1.98 \\
 PCWNet~\cite{pcwnet}       & \bf{1.37} & 3.16      & 1.67 \\
 GMStereo~\cite{unimatch}   & 1.49      & 3.14   & 1.77 \\
 ACVNet~\cite{acvnet}       & \bf{1.37} & 3.07      & \ul{1.65} \\
 LEAStereo~\cite{leastereo} & 1.40      & 2.91      & \ul{1.65} \\
 CREStreo~\cite{crestereo}  & 1.45      & \ul{2.86} & 1.69 \\
 \textbf{\ours}      & 1.38      & \bf{2.65} & \bf{1.59} \\
 \bottomrule
 \end{tabular}
 %} 
 \normalsize
 \caption{\textbf{Evaluation on the KITTI 2015 stereo benchmark} with the percentage of outliers (\ie, error above 3 pixels) for background (D1-bg), foreground (D1-fg) and all (D1-all) pixels.}
 \label{tab:kitti}
\end{table}

On Middlebury (Table~\ref{tab:middlebury2}), \ours obtains the lowest average error on 6 out of 15 sequences, in spite of using a generic patch-based transformer without any of the usual apparatus for stereo matching (\eg cost-volume, coarse-to-scale processing, iterative refinement).
However, in average, we obtain a worse error due to the fact that \ours produces really large errors for a few sequences like Hoops or ClassE.
In fact, these errors correspond to cases with large maximum disparities (based on the maximum threshold value applied before evaluation), which is harmful for our simple tiling-based inference approach.
This effect is visible in the prediction of the bottom example of Figure~\ref{fig:examples}
where one can observe tiling artefacts, \eg next to the stair pillars.
In general, however, our method remains accurate, especially on thin structures like the pins on the map or the radius of the bicycle wheels in Figure~\ref{fig:examples}.

For KITTI 2015 (Table~\ref{tab:kitti}), we finetune \ours on $1216{\times}352$ crops from KITTI 2012~\cite{kitti} and 2015~\cite{kitti15} for 20 epochs. 
\ours performs the best on the main D1-all metrics (outliers ratio at a 3px error threshold), with the best value also on foreground pixels, and at 0.01\% of the best methods on background pixels.

For ETH3D, we use a Laplacian loss without bounds as it is limited to small disparities, \ie, with parameterization $d_i = e^{d'_i}$ and weights $e^{-3 d'_i}$ for tiling. \ours sets a new state of the art for the ratio of pixels with an error over 0.5px (bad@0.5) and performs on par with CREStereo~\cite{crestereo} for bad@1.0 and the average error, see Table~\ref{tab:eth3d}. It outperforms recent approaches like GMStereo~\cite{unimatch}, RAFT-Stereo~\cite{raftstereo}, DIP-Stereo~\cite{dip} or HITNet~\cite{hitnet} by a large margin, \eg the bad@0.5 for non-occluded pixels is improved by 3\% or more.

\begin{table}[]
    \centering
    \resizebox{\linewidth}{!}{
    \begin{tabular}{lcccccc}
    \toprule
    \multirow{2}{*}{Method} & \multicolumn{2}{c}{\small bad@0.5 (\%)$\downarrow$} & \multicolumn{2}{c}{\small bad@1.0 (\%)$\downarrow$} & \multicolumn{2}{c}{\small avg err (px)$\downarrow$} \\
              & noc & all & noc & all & noc & all \\
    \cmidrule(lr){1-1} \cmidrule(lr){2-3} \cmidrule(lr){4-5} \cmidrule(lr){6-7}
    AdaStereo~\cite{adastereo} & 10.22 & 10.85 & 3.09 & 3.34 & 0.24 & 0.25 \\    
    HITNet~\cite{hitnet} & 7.89 & 8.41 & 2.79 & 3.11 & 0.20 & 0.22 \\    
    RAFT-Stereo~\cite{raftstereo} & 7.04 & 7.33 & 2.44 & 2.60 & 0.18 & 0.19 \\    
    DIP-Stereo~\cite{dip} & 6.74 & 6.99 & 1.97 & 2.12 & 0.18 & 0.20 \\    
    GMStereo~\cite{unimatch} & 5.94 & 6.44 & 1.83 & 2.07 & 0.19 & 0.21 \\
    CREStereo~\cite{crestereo} & \ul{3.58} & \ul{3.75} & \bf{0.98} & \bf{1.09} & \bf{0.13} & \bf{0.14} \\
     \textbf{\ours}   & \bf{3.27} & \bf{3.51} & \ul{0.99} & \ul{1.14} & \ul{0.14} & \ul{0.15} \\
    \bottomrule
    \end{tabular}
    }
    \caption{\textbf{Evaluation on ETH3D} with the percentage of pixels with an error over 0.5px (bad@0.5), over 1px (bad@1.0) and the average error over non-occluded (noc) or all pixels.}
    \label{tab:eth3d}
\end{table}

Finally, we report results on the recent Spring benchmark in Table~\ref{tab:springstereo} where our model is finetuned for 8 epochs on its training set. \ours outperforms the leading methods on all metrics with a large margin, \ie, the main bad@1 metric is reduced from $15\%$ to $7\%$ and the absolute error from $1.5$ to $0.5$px.

\begin{table}[]
\resizebox{\linewidth}{!}{
\begin{tabular}{lccccc}
\toprule
Method & {\small 1px$\downarrow$} & {\small 1px s0-10$\downarrow$} & {\small 1px s10-40$\downarrow$} & {\small 1px s40+$\downarrow$} & {\small Abs$\downarrow$} \\ 
\midrule
RAFT-Stereo~\cite{raftstereo}$^\ddagger$ & 15.273 & 22.588 & \ul{10.018} & \ul{17.086} & 3.025 \\
AVC-Net~\cite{acvnet}$^\ddagger$ & \ul{14.772} & \ul{18.386} & 11.346 & 18.145 & \ul{1.516} \\
\bf{\ours} & \bf{7.135} & \bf{2.934} & \bf{7.757} & \bf{13.247} & \bf{0.471} \\
\bottomrule
\end{tabular}
}
\caption{\textbf{Evaluation of \ours on the Spring benchmark} with the percentage of outliers (error over 1px) over all pixels, or over pixels with disparities in [0,10] (s0-10), in [10,40] (s10-40) and over 40 pixels (s40+), as well as the average absolute error (Abs). $^\ddagger$ means methods submitted by the leaderboard's authors.}
\label{tab:springstereo}
\end{table}

\PAR{\oursflow \vs the state of the art} 
We compare \oursflow to the state of the art on the official leaderboards of MPI-Sintel~\cite{sintel}, KITTI 2015~\cite{kitti15} and Spring~\cite{spring}.

\begin{figure}
\newcommand{\fexlength}{0.245\linewidth}
\resizebox{\linewidth}{!}{
\begin{tabular}{c@{ }c@{ }c@{ }c@{ }c}
{\small First image} & {\small Ground truth} & 
{\small GMFlow+~\cite{unimatch}} & {\small \bf{\oursflow}} \\
\includegraphics[width=\fexlength]{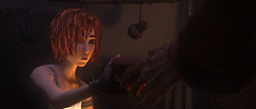} &
\includegraphics[width=\fexlength]{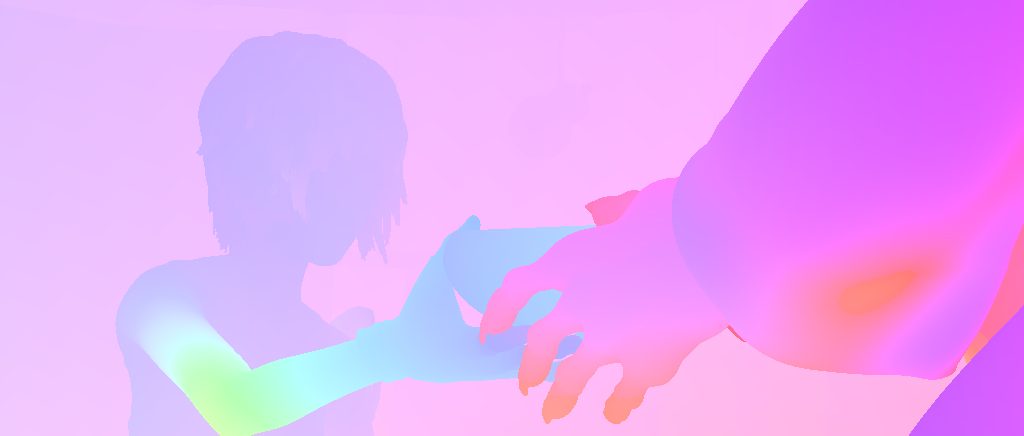} &
\includegraphics[width=\fexlength]{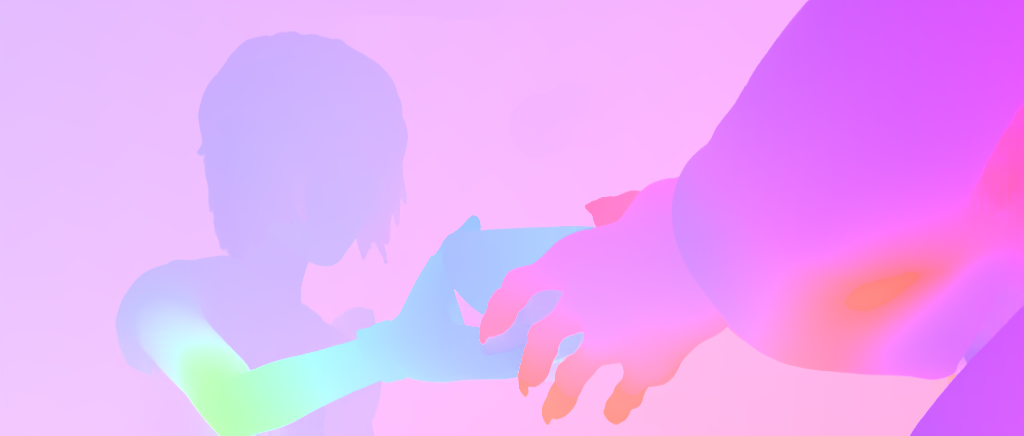} &
\includegraphics[width=\fexlength]{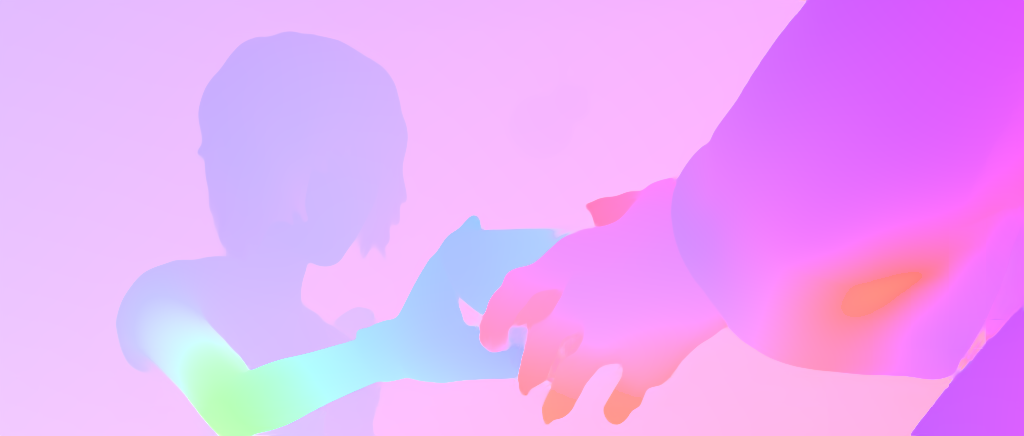} 
 \\[-0.05cm]
\includegraphics[width=\fexlength]{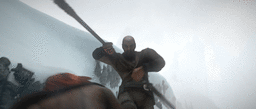} &
\includegraphics[width=\fexlength]{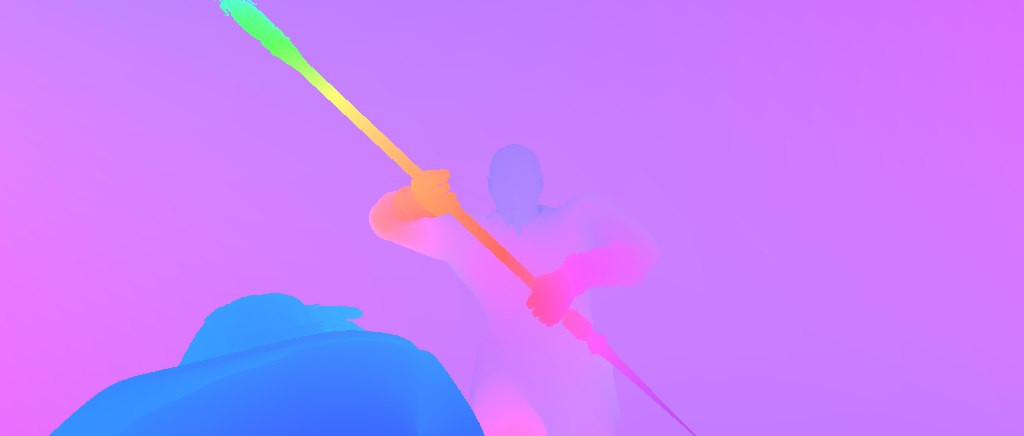} &
\includegraphics[width=\fexlength]{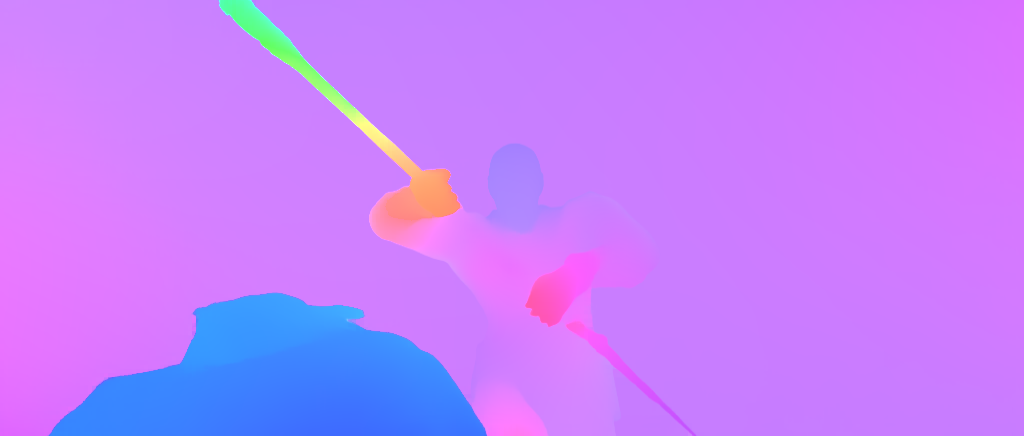} &
\includegraphics[width=\fexlength]{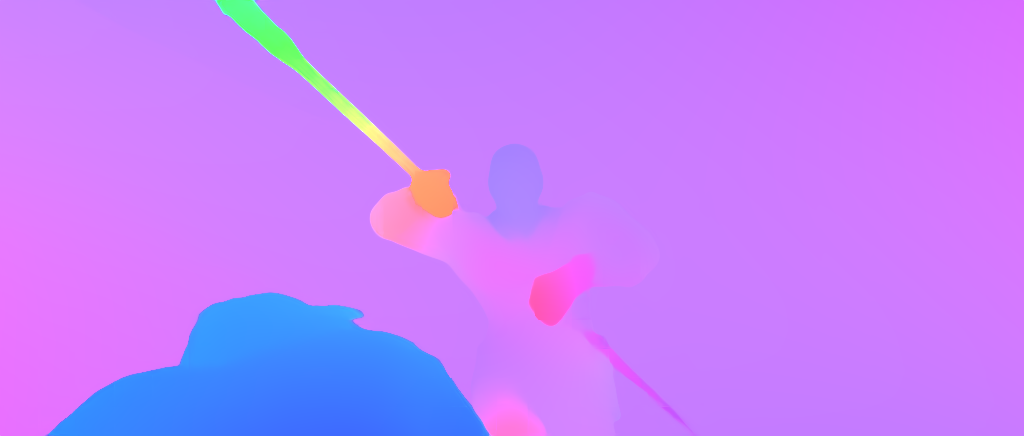}
\end{tabular}
}
\\[-0.25cm]
\caption{\textbf{Two examples from the MPI-Sintel test set} with from left to right: the first image, the ground truth, 
GMFlow+~\cite{unimatch} and \oursflow.}
\label{fig:flowexamples}
\end{figure}

On MPI-Sintel (Table~\ref{tab:sintel}), \oursflow performs better than RAFT~\cite{raft} which include many specialized refinement steps and use previous flow estimation as initialization. We rank second on the clean rendering and perform competitively on the final rendering, on par with most recent approaches such as GMFlow+~\cite{unimatch}, SKFlow~\cite{skflow} or FlowFormer~\cite{flowformer}. Figure~\ref{fig:flowexamples} shows some visualizations of flow prediction.

\begin{table}[]
 \centering
 %\small
 %\resizebox{\linewidth}{!}{
 \begin{tabular}{lcc}
 \toprule
 Method & clean$\downarrow$ & final$\downarrow$ \\
 \midrule
PWC-Net+~\cite{pwcnetplus}   & 3.45 & 4.60 \\
RAFT$^\dagger$~\cite{raft}   & 1.61 & 2.86 \\
 CRAFT$^\dagger$~\cite{craft} & 1.44 & 2.42 \\
 FlowFormer~\cite{flowformer} & 1.20 & \bf{2.12} \\
 SKFlow~\cite{skflow}     & 1.30 & 2.26 \\
 GMFlow+~\cite{unimatch}  & \bf{1.03} &  \bf{2.12} \\
 \textbf{\oursflow}      & \ul{1.09} &  2.44 \\
 \bottomrule
 \end{tabular}
 %} 
 \normalsize
 \caption{\textbf{Evaluation on the MPI-Sintel benchmark} with the EPE~($\downarrow$) on the clean and final renderings. $^{\dagger}$ means that the flow prediction from the previous frames is used as initialization.}
 \label{tab:sintel}
\end{table}

For KITTI 2015 (Table~\ref{tab:kittiflow}), we finetuned the model on the training set from KITTI 2012 and 2015 for 150 epochs on crops of size $1216{\times}352$. \oursflow performs best on the main F1-all metrics, \ie, the percentage of outliers, with a large margin: the F1-all is reduced from 4.49\% to 3.64\% compared to GMFlow+~\cite{unimatch}. This gap mainly comes from the background pixels, while we perform on par with the best methods on foreground pixels.

Finally, on Spring, for which we finetune the model on its training set for 12 epochs, we obtain state-of-the-art performance, see Table~\ref{tab:springflow}.
We obtain an EPE of 0.50, compared to 0.64 for the second best method, with an outlier ratio reduced for all flow norm ranges.

\PAR{Limitations}
The tiling-based inference strategy may prevent an accurate estimate in case of extremely large disparity or flow, where the corresponding pixels can be outside of the tile of the second image.
A tiling strategy smarter than taking the same cropping coordinates in a pair of images could be considered. 

\begin{table}[]
\centering
\begin{tabular}{lccc} 
    \toprule
    Method & {\small Fl-bg$\downarrow$} & {\small Fl-fg$\downarrow$} & {\small Fl-all$\downarrow$} \\
    \midrule
    PWC-Net+~\cite{pwcnetplus} & 7.69 & 7.88 & 7.72 \\
    RAFT$^\dagger$~\cite{raft}   & 4.74 & 6.87 & 5.10 \\
    CRAFT$^\dagger$~\cite{craft}   & 4.58 & \ul{5.85} & 4.79 \\
     FlowFormer~\cite{flowformer} & 4.37 & 6.18 & 4.68 \\
    GMFlow+~\cite{unimatch} & \ul{4.27} & \bf{5.60} & \ul{4.49} \\ 
    \bf{\oursflow} & \bf{3.18} & 5.94 & \bf{3.64} \\
    \bottomrule
\end{tabular}
\caption{\textbf{Evaluation of \oursflow on the KITTI 2015 benchmark} with the percentage of outliers for background (F1-bg), foreground (F1-fg) and all (F1-all) pxiels. $^{\dagger}$ means that the flow prediction from the previous frames is used as initialization.}
\label{tab:kittiflow}
\end{table}

\begin{table}
\centering
\resizebox{\linewidth}{!}{
\begin{tabular}{lccccc}
\toprule
Method & {\small 1px$\downarrow$} & {\small 1px s0-10$\downarrow$} & {\small 1px s10-40$\downarrow$} & {\small 1px s40+$\downarrow$} & {\small EPE$\downarrow$} \\ 
\midrule
FlowFormer~\cite{flowformer}$^\ddagger$ & 6.510 & 3.381 & 5.530 & \ul{35.344} & 0.723 \\
MS-Raft+~\cite{msraftplus}$^\ddagger$ & \ul{5.724} & \ul{2.055} & \ul{5.022} & 38.315 & \ul{0.643} \\
\bf{\oursflow} & \bf{4.565} & \bf{1.225} & \bf{4.332} & \bf{33.134} & \bf{0.498} \\
\bottomrule
\end{tabular}
}
\caption{\textbf{Evaluation of \oursflow on the Spring benchmark} with the number of outliers (error over 1px) over all pixels, or over pixels with flow norm in [0,10] (s0-10), in [10,40] (s10-40) and over 40 pixels (s40+) as well as the endpoint error (EPE). $^\ddagger$ means methods submitted by the leaderboard's authors.}
\label{tab:springflow}
\end{table}

%%%%%%%%%% Conclusion
\section{Conclusion}
\label{sec:conclusion}

For the first time, we have shown that large-scale pre-training can be successful for dense geometric tasks, 
thanks to a well-adapted pretext task and real-world data at scale. 
This enables to reach state-of-the-art performance with a ViT-based architecture without using task-specific designs, 
and thereby opening novel routes to tackle these problems, and new avenues towards more universal vision models.

%%%%%%%%% REFERENCES
{\small
\bibliographystyle{ieee_fullname}
\bibliography{biblio}
}

%%%%%%%%%%%%%%%%%%%%%%%%%%%%%%%%%%%%%%%%%%%%%%%%%%%%%%%%%%% appendix
\clearpage
%\vspace{2cm}

\newcommand{\completionexample}[1]{\includegraphics[width=0.195\linewidth]{fig/completion_examples/#1_ref.png} & \includegraphics[width=0.195\linewidth]{fig/completion_examples/#1_maskedtgt.png} & \includegraphics[width=0.195\linewidth]{fig/completion_examples/#1_pred_neurips.png} & \includegraphics[width=0.195\linewidth]{fig/completion_examples/#1_pred_v2.png} & \includegraphics[width=0.195\linewidth]{fig/completion_examples/#1_tgt.png} \\[-0.05cm] }

\twocolumn[{%
\renewcommand\twocolumn[1][]{#1}%
\appendix
\begin{center}
    \captionsetup{type=figure}
\resizebox{\linewidth}{!}{
\begin{tabular}{@{}c@{ }c@{ }c@{ }c@{ }c@{}}
Reference image & Masked target & CroCo~\cite{croco} & \textbf{\croconew} & Target image \\
\completionexample{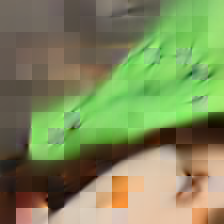}
\completionexample{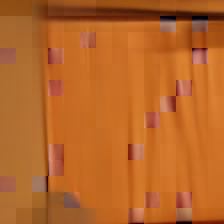}
\completionexample{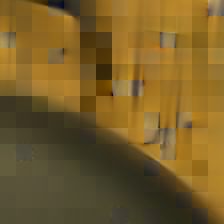}
\end{tabular}
}
\\[-0.4cm]
\captionof{figure}{\textbf{Cross-view reconstruction examples} (pre-training pretext task) on scenes unseen during pretraining for the original CroCo~\cite{croco} and with our improvements. The images come from the Middlebury stereo benchmark~\cite{middlebury}.
}
\label{fig:reconstructions1}

\vspace{0.8cm}

\end{center}
}]

{\huge \bf{Appendix} \vspace{0.5cm}}

In this appendix, we first provide visualizations of the capabilities of \croconew on the pretext task of cross-view completion (Section~\ref{app:crocovis}).
We then present additional experimental results in Section~\ref{app:morexp}, including in particular (a) the impact of pre-training, (b) the runtime of our model and (c) an analysis of the probabilistic distributions regressed by our \ours model for the stereo matching task. We finally detail our training setup and the dataset splits. (Section~\ref{app:details}).

\begin{figure*}[t]
\centering
\resizebox{\linewidth}{!}{
\begin{tabular}{@{}c@{ }c@{ }c@{ }c@{ }c@{}}
Reference image & Masked target & CroCo~\cite{croco} & \textbf{\croconew} & Target image \\
\completionexample{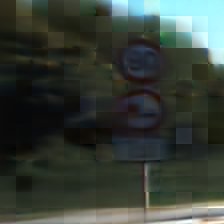}
\completionexample{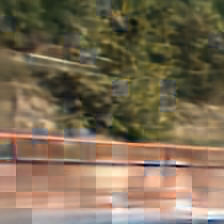}
\completionexample{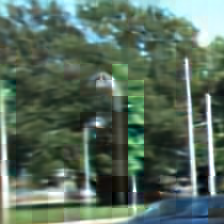}
\end{tabular}
}
\\[-0.4cm]
    \caption{\textbf{Cross-view reconstruction examples} (pre-training pretext task) on scenes unseen during pre-training for the original CroCo~\cite{croco} and with our improvements. The images come from the stereo benchmark of KITTI~\cite{kitti}.
    }
    \label{fig:reconstructions2}
\end{figure*}

\section{Cross-view completion examples}
\label{app:crocovis}

To qualitatively evaluate the impact of \croconew, \ie, of the improvements that we propose on top of the CroCo~\cite{croco} pre-training, we show several examples of cross-view completions on real-world scenes, coming either from Middlebury v3~\cite{middlebury} in Figure~\ref{fig:reconstructions1} or KITTI~\cite{kitti15} in Figure~\ref{fig:reconstructions2}.
Note that these methods, as MAE~\cite{mae}, regress pixel values that are normalized according to the mean and standard deviation inside each patch, we thus apply the inverse transform for display: this means that the overall color of each patch will be correct, as it comes from the ground-truth values.
While the most important measure of performance of these models is their transfer to downstream tasks, as explored in the main paper, a qualitative observation of the fact that our improved method is better at solving the pretext task is noteworthy.
We clearly observe that the reconstructions from the original CroCo~\cite{croco} tend to be quite blurry in many areas, which might come from the fact that it relies on a smaller model and was pre-trained only on synthetic data from indoor environments, while details are impressively preserved thanks to our improvements. 
In Figure~\ref{fig:reconstructions1}, note how the lines and the eyes are well reconstructed in the first row, or the roads on the maps of the third row, despite the high masking ratio that is applied to the masked image (90\%).
Similarly, the text is clearly readable on the first row of Figure~\ref{fig:reconstructions2}. Some predictions by our model have some blur (\eg left of the first and thirds rows of Figure~\ref{fig:reconstructions1}), which makes sense because these parts are not visible in the reference image.

\section{Further experimental results}
\label{app:morexp}

\begin{table*}
\centering
%\resizebox{\linewidth}{!}{
\begin{tabular}{llp{0cm}rrrrp{0cm}rrrr}
     \toprule
& \multirow{2}{*}{Network initialization} & & \multicolumn{4}{c}{Stereo (bad@1.0px$\downarrow$)} & & \multicolumn{4}{c}{Flow (EPE$\downarrow$)}  \\
& & & \multicolumn{1}{c}{\small Md} & \multicolumn{1}{c}{\small ETH} & \multicolumn{1}{c}{\small SF(c)} & \multicolumn{1}{c}{\small SF(f)} & & \multicolumn{1}{c}{\small FT(c)} & \multicolumn{1}{c}{\small FT(f)} & \multicolumn{1}{c}{\small Si.(c)} & \multicolumn{1}{c}{\small Si.(f)}  \\
\cmidrule(lr){1-2} \cmidrule(lr){4-7} \cmidrule(lr){9-12}
\multicolumn{12}{l}{\emph{RoPE positional embedding, ViT-L encoder, Base decoder, 2M Habitat + 5.3M real pre-training pairs}} \\
& \bf{\croconew pre-training} && \bf{15.5} &  \bf{0.38} &   \bf{5.0} &   \bf{5.3} &       &\bf{2.85} &  \bf{2.45} &  \bf{1.43} &  \bf{1.99} \\
& random init.       &&   43.4 &  1.06 &  11.0 &  11.2     &    &  10.53 & 10.57 &  4.84 &  5.49 \\
\cmidrule(lr){1-2} \cmidrule(lr){4-7} \cmidrule(lr){9-12} 
\multicolumn{12}{l}{\emph{cosine positional embedding, ViT-B encoder, Small decoder, 2M Habitat (synthetic only) pre-training pairs}} \\
& CroCo~\cite{croco} pre-training && \bf{26.3} &  1.82 &   \bf{6.7} &   \bf{7.0} &    &   \bf{3.89} &  \bf{3.56} &  \bf{2.07} &  \bf{2.57}     \\
& MAE~\cite{mae} (ImageNet) pre-training (encoder only)
&& 35.8 &  \bf{1.68} &   8.6 &   8.8 &    &   5.13 &  4.83 &  2.92 &  3.82          \\
& random init.   && 87.5 &  5.42 &  24.6 &  24.6 &    &  14.28 & 14.31 &  8.99 &  9.76     \\
\bottomrule
\end{tabular}
%} 
\vspace{-0.3cm}
\caption{\textbf{Impact of pre-training.} We compare the performance of our final model (first row) with improved cross-view completion pre-training to a randomly initialized version (second row). To compare to MAE~\cite{mae}, that is pre-trained on ImageNet~\cite{imagenet}, and which is based on cosine positional embeddings, we make the comparison with the original CroCo in the bottom rows.}
\label{tab:pretrain}
\end{table*}

\begin{table*}
\centering
%\resizebox{\linewidth}{!}{
\begin{tabular}{cp{0cm}rrrrp{0cm}rrrr}
     \toprule
Masking & & \multicolumn{4}{c}{Stereo (bad@1.0px$\downarrow$)} & & \multicolumn{4}{c}{Flow (EPE$\downarrow$)}  \\
ratio & & \multicolumn{1}{c}{\small Md} & \multicolumn{1}{c}{\small ETH} & \multicolumn{1}{c}{\small SF(c)} & \multicolumn{1}{c}{\small SF(f)} & & \multicolumn{1}{c}{\small FT(c)} & \multicolumn{1}{c}{\small FT(f)} & \multicolumn{1}{c}{\small Si.(c)} & \multicolumn{1}{c}{\small Si.(f)}  \\
\cmidrule(lr){1-1} \cmidrule(lr){3-6} \cmidrule(lr){8-11}
80\% & &   32.5 &  1.96 &   7.3 &   7.5 &    &   4.29 &  4.06 &  2.06 &  2.71 \\
85\% & &   59.2 &  1.15 &   8.7 &   9.0 &    &   3.48 &  3.08 &  1.99 &  2.41 \\
\bf{90\%} & &  \bf{20.7} &  \bf{0.82} &   \bf{5.8} &   \bf{6.1} &    &   \bf{3.35} &  \bf{2.94} &  \bf{1.76} &  \bf{2.30} \\
\bottomrule
\end{tabular}
%} 
\vspace{-0.3cm}
\caption{\textbf{Impact of the pre-training masking ratio} for a model with RoPE positional embeddings, a ViT-B encoder, a Small decoder, pre-trained on 2M Habitat + 5.3M real pairs.}
\label{tab:maskratio}
\end{table*}

\subsection{Impact of pre-training} 

In Table~\ref{tab:pretrain}, we measure the impact of the pre-training on the downstream performance when the model is finetuned for stereo matching or optical flow. The first two rows compare our model, using our improved cross-view completion pre-training \vs a random initialization. We observe a clear gain of performance, \eg on the FlyingThings flow test set in the final rendering with an EPE of 2.45 pixels with pre-training \vs 10.57 without it, or on the Middlebury v3 stereo validation set with a bad@1.0px of 15.5\% with pre-training \vs 43.4\% without it.

We are not aware of any other pre-training strategy, other than cross-view completion, that readily includes a binocular decoder or architecture. While it is still possible to initialize part of the layers using other pre-training strategies, this means that some important parts of the network are still being initialized at random. Nevertheless, to compare to other pre-training strategies, we consider MAE~\cite{mae} pre-trained on ImageNet~\cite{imagenet}, thus with a cosine positional embedding, a ViT-Base encoder, and with a Small decoder that is randomly initialized. We compare that to the original CroCo~\cite{croco} pre-trained on synthetic data only. We observe that CroCo pre-training obtains the lowest errors, significantly outperforming the MAE pre-training and the random initialization.

Interestingly, the performance of this smaller model is also significantly better than the large one without pre-training. This again highlights the importance of the pre-training with such generic architecture.

\vpara{Masking ratio.} CroCo~\cite{croco} finds that using a masking ratio of 90\% performs best for cross-view completion on their synthetic data. This is higher than the 75\% masking ratio of MAE~\cite{mae}, as the unmasked reference view of the same scene adds redundancy. 
A question is whether this masking ratio of 90\% that has been found optimal on synthetic data generalizes to real data.
Table~\ref{tab:maskratio} reports the performance on stereo and flow downstream tasks for a masking ratio of 80\%, 85\% and 90\%.
We find that a masking ratio of 90\% performs best also in the case of using real data.

\subsection{Smaller training data}

\begin{table}
\centering
\begin{tabular}{lcc}
\toprule
 \multirow{2}{*}{Method} & \multicolumn{2}{c}{MPI-Sintel($\downarrow$)} \\
 & clean & final \\
\cmidrule(lr){1-1} \cmidrule{2-3}
LiteFlowNet2~\cite{liteflownet2} & 2.24 & 3.78 \\
FM-RAFT~\cite{fmraft} & 1.29 & 2.95 \\
FlowFormer~\cite{flowformer} & \bf{1.01} & \bf{2.40} \\ 
RAFT~\cite{raft} before refinement        & 4.04 & 5.45 \\
RAFT~\cite{raft} & 1.41 & 2.69 \\ 
GMFlow~\cite{unimatch} before refinement  & 1.31 & 2.96 \\
GMFlow~\cite{unimatch} & \ul{1.08} & \ul{2.48} \\
\textbf{\oursflow} & 1.28 & 2.58 \\
\bottomrule
\end{tabular} \\[-0.3cm]
\caption{\textbf{Optical flow results when training on FlyingChairs and FlyingThings only.} We report the EPE on MPI-Sintel training set (clean or final rendering). Numbers for the first three rows come from~\cite{flowformer}, numbers for RAFT and GMFlow (before and after refinement) from~\cite{unimatch}.}
\label{tab:flying}
\end{table}

Most optical flow methods also report the performance on the MPI-Sintel training set when training on FlyingChairs and FlyingThings only. We report these values in Table~\ref{tab:flying}. For RAFT~\cite{gmflow} and GMFlow~\cite{unimatch}, we report the numbers before and after using iterative refinement procedures. Interestingly, \oursflow performs better than these two methods before their refinement. Overall, our ranking is similar to the ones on the MPI-Sintel test set where we use more training data. This indicates that our finetuning on geometric downstream tasks do not necessarily need large-scale training data, despite the size of our architecture.

\subsection{Runtime and tiling}

\begin{table}[]
\centering
\resizebox{\linewidth}{!}{
\begin{tabular}{cccrr@{ }l}
\toprule
Pos. & Encoder & Decoder & runtime & \multicolumn{2}{l}{\#Parameters} \\ 
\midrule
cosine & ViT-B & Small & 25ms & 139.4M & (85.6M+34.0M+19.7M) \\
RoPE & ViT-B & Small & 26ms & 139.4M & (85.6M+34.0M+19.7M) \\
RoPE & ViT-B & Base & 29ms & 219.7M & (85.6M+114.0M+20.1M) \\
RoPE & ViT-L & Base & 53ms & 437.4M & (303.1M+114.2M+20.1M) \\
\bottomrule
\end{tabular}
}
\\[-0.3cm]
\caption{\textbf{Runtime and number of parameters.} Runtime is measured for a single tile of size $704{\times}352$, on a NVIDIA A100 GPU. For the number of parameters we report in parenthesis the numbers for the encoder, the decoder and the DPT head separately.}
\label{tab:runtime}
\end{table}

\vpara{Runtime.} In Table~\ref{tab:runtime}, we report the runtime for different sizes of our model. On one single tile of the same size as training for stereo, \ie, $704{\times}352$, on a NVIDIA A100 GPU. Our method remains relatively fast on one tile, in the order of a few tens of milliseconds.

\input{plot/tiling/styles}
\input{plot/tiling/plots}
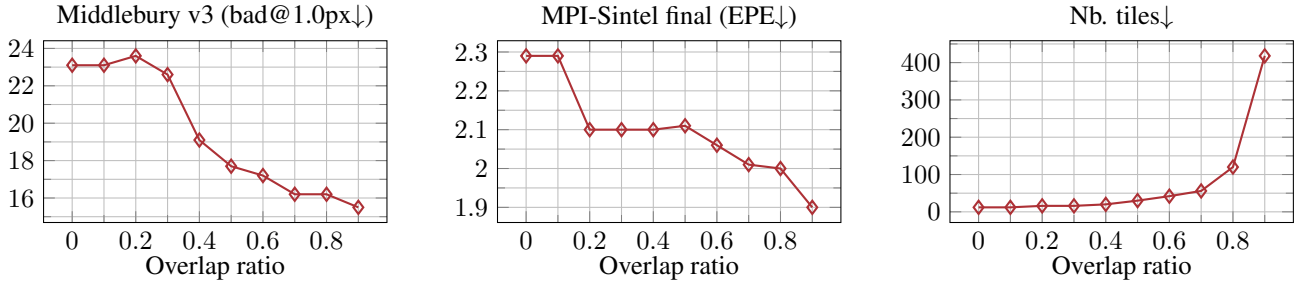
\begin{figure*}[!th]
\centering
\begin{subfigure}{.32\linewidth}

\begin{tikzpicture}
\begin{axis}[%
  width=1.1\linewidth,
  height=4cm, 
   xlabel={\normalsize Overlap ratio},
   xlabel style={yshift=4pt},
  minor tick num=1,
  ticklabel style = {font=\normalsize},
  legend pos=north east,
    % legend pos=outer north east,
    % xmode=log,
    title={\normalsize Middlebury v3 (bad@1.0px$\downarrow$)},
            title style={yshift=-5pt},
  ]
\input{plot/tiling/data}
    \addplot[fin]      table[x=overlap,  y=bad_md] \map;
    %\addplot[crcap]      table[x=overlap,  y=bad_sfclean] \map;
    %\addplot[crcap]      table[x=cf_epoch,  y=cf_pampjpe] \map;
    % \addplot[maep]      table[x=ca_epoch,  y=ca_pampjpe]  \map; 
     %\addplot[crca]      table[x=inet_epoch,  y=inet_pampjpe]  \map; 
      %\addplot[crcb]      table[x=dino_epoch,  y=dino_pampjpe]  \map; 
      %\addplot[crca]      table[x=mae_epoch,  y=mae_pampjpe]  \map; 
     %\addplot[crcd]      table[x=croco_epoch,  y=croco_pampjpe]  \map; 
      %\addplot[crce]      table[x=nomc_epoch,  y=nomc_pampjpe]  \map; 
      %\addplot[gvert] coordinates {(0, 58.93) (50, 58.93)};
\end{axis}
\end{tikzpicture}
\end{subfigure}
\hfill
\begin{subfigure}{.32\linewidth}
%\plotSf

\begin{tikzpicture}
\begin{axis}[%
  width=1.1\linewidth,
  height=4cm, 
   xlabel={\normalsize Overlap ratio},
   xlabel style={yshift=4pt},
  minor tick num=1,
  ticklabel style = {font=\normalsize},
  legend pos=north east,
    % legend pos=outer north east,
    % xmode=log,
    title={\normalsize MPI-Sintel final (EPE$\downarrow$)},
            title style={yshift=-5pt},
  ]
\input{plot/tiling/data}
    \addplot[fin]      table[x=overlap,  y=epe_sintelf] \map;
    %\addplot[crcap]      table[x=overlap,  y=bad_sfclean] \map;
    %\addplot[crcap]      table[x=cf_epoch,  y=cf_pampjpe] \map;
    % \addplot[maep]      table[x=ca_epoch,  y=ca_pampjpe]  \map; 
     %\addplot[crca]      table[x=inet_epoch,  y=inet_pampjpe]  \map; 
      %\addplot[crcb]      table[x=dino_epoch,  y=dino_pampjpe]  \map; 
      %\addplot[crca]      table[x=mae_epoch,  y=mae_pampjpe]  \map; 
     %\addplot[crcd]      table[x=croco_epoch,  y=croco_pampjpe]  \map; 
      %\addplot[crce]      table[x=nomc_epoch,  y=nomc_pampjpe]  \map; 
      %\addplot[gvert] coordinates {(0, 58.93) (50, 58.93)};
\end{axis}
\end{tikzpicture}
\end{subfigure}
\hfill
\begin{subfigure}{.32\linewidth}

\begin{tikzpicture}
\begin{axis}[%
  width=1.1\linewidth,
  height=4cm, 
   xlabel={\normalsize Overlap ratio},
   xlabel style={yshift=4pt},
  minor tick num=1,
  ticklabel style = {font=\normalsize},
  legend pos=north east,
    % legend pos=outer north east,
    % xmode=log,
    title={\normalsize Nb. tiles$\downarrow$},
            title style={yshift=-5pt},
  ]
\input{plot/tiling/data}
    \addplot[fin]      table[x=overlap,  y=nbtiles] \map;
\end{axis}
\end{tikzpicture}
\end{subfigure}
\\[-0.4cm]
\caption{\textbf{Impact of the overlap ratio between tiles during inference.} We plot the stereo performance (bad@1.0px in \%) on Middlebury v3 (left) and the flow performance on MPI-Sintel in its final rendering (middle) when varying the overlap ratio during inference with tiling. We also plot the number of tiles it represents for a $1920{\times}1080$ image (right), which is proportional to the total runtime, for a crop size of $704{\times}352$ as \ours. 
}
\label{fig:overlap}
\end{figure*}

\vpara{Number of parameters.} We also report the number of trainable parameters in Table~\ref{tab:runtime}. 
This number of parameters is one order of magnitude higher than most existing stereo and flow methods.
We did not study how this number of parameters could be reduced, and we also do not claim that our models are better than existing work for a fixed computational budget. 
Indeed, task-specific approaches have the advantage of being more sample efficient, \ie, requiring less data, by leveraging prior knowledge about the task. They also have the drawback of not being readily compatible with large-scale training on unlabeled data, because of task-dependent components, which limits the use of large generic models.
Existing methods cannot be scale up to a larger number of parameters easily, as training large models requires lot of data. In the case of stereo and optical flow, for which labeled data is limited, 
this means using self-supervised learning, which cannot be straightforwardly applied for models that involve task-specific designs like cost volumes, image warping, \etc. 
Thus, our contribution and our aim in this work is to show that pre-training large, generic architectures and finetuning them for stereo matching and optical flow is a valid path forward.

\vpara{Impact of the overlap ratio during tiling.} In Figure~\ref{fig:overlap}, we report the performance and the number of tiles for a Full HD image ($1920{\times}1080$) when varying the overlap ratio during inference with the tiling strategy. While the performance improves with a higher overlap ratio, the number of tiles can rapidly explodes. With an overlap around 0.5 or 0.7, performance is quite close to the one obtained with 0.9 while the number of tiles remains reasonable. This may thus be the best trade-off in practical scenarios where inference time has to stay small.

\subsection{Laplacian-based loss}
\label{appsub:loss}

\begin{figure}
\centering
\newcommand{\confexample}[1]{\includegraphics[width=0.33\linewidth]{fig/confidence/#1_img1.png} & \includegraphics[width=0.33\linewidth]{fig/confidence/#1_err.png} & \includegraphics[width=0.33\linewidth]{fig/confidence/#1_conf.png} \\[-0.05cm] }
\resizebox{\linewidth}{!}{
\begin{tabular}{@{}c@{ }c@{ }c@{}}
First image & Prediction error & Uncertainty \\ 
\confexample{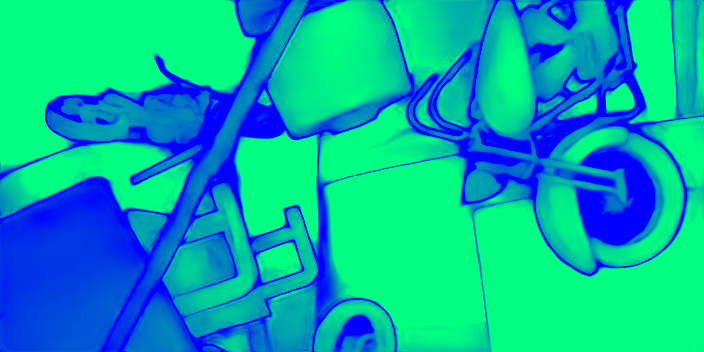}
\confexample{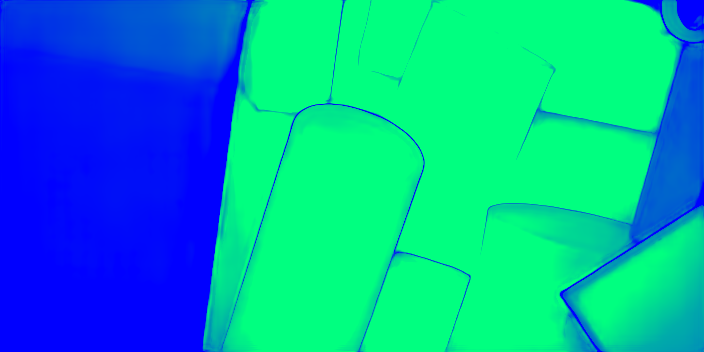}
\confexample{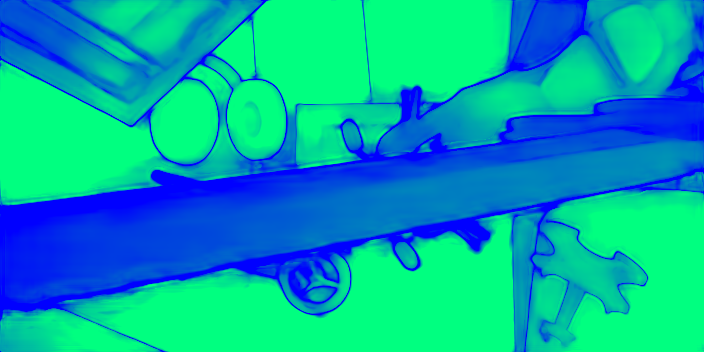}
\confexample{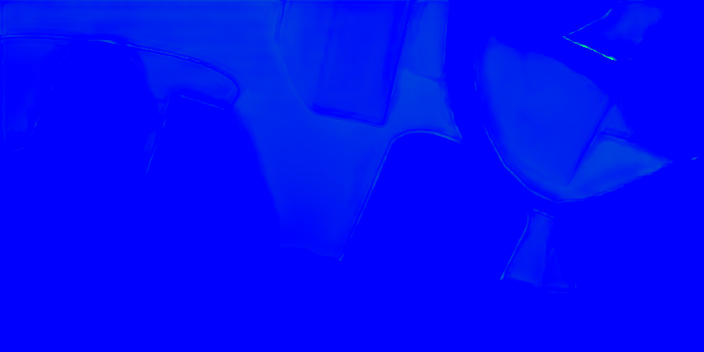}
\end{tabular}
}

\vspace{-0.3cm}

\caption{\textbf{Visualization of the uncertainty predicted by \ours} on a few examples from the SceneFlow test set. The first column shows the first image, the second column shows the error of the prediction clamped within the segment $[0,10]$, the third column shows the logarithm of the predicted scale of the Laplacian distribution output by the model: green colors denote confident areas while blue colors denote uncertain areas.}
\label{fig:confidence}    
\end{figure}

For flow and stereo, we regress a Laplacian distribution: the location parameter corresponds to the disparity or flow prediction, while the scale parameter could be seen as a measure of uncertainty. We thus denote here by `uncertainty' the logarithm of the predicted scale of the Laplacian distribution that our downstream model outputs, \ie, $log(d_i)$ from Equation~\ref{eq:loss}.

\begin{figure*}
\centering
\includegraphics[width=0.4\linewidth]{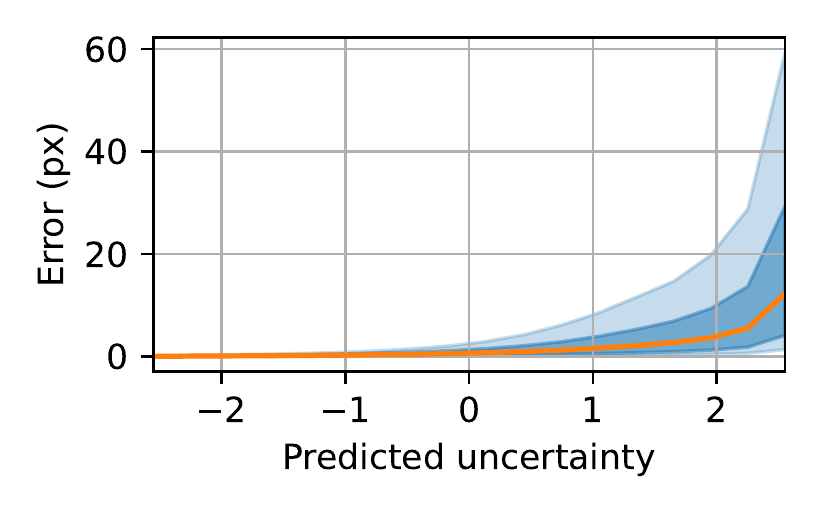} 
\quad \quad 
\includegraphics[width=0.4\linewidth]{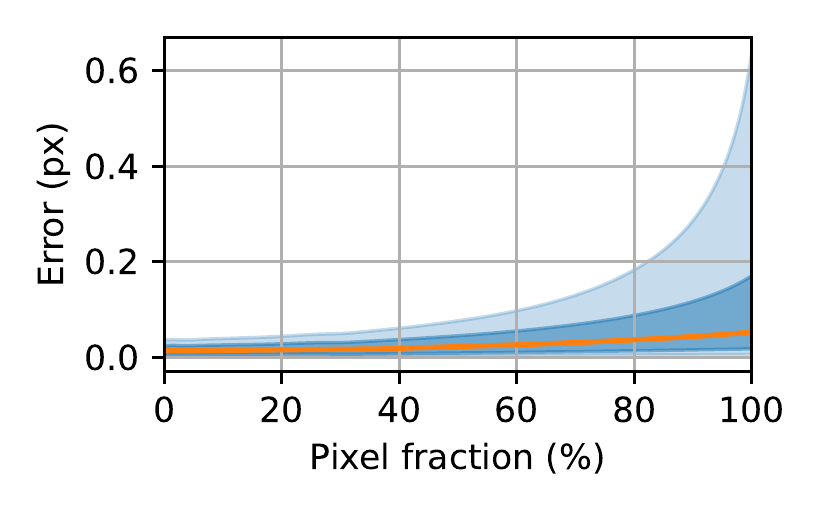} \\[-0.6cm]
\caption{\textbf{Statistics on the uncertainty predicted by \ours.} We subsample 1000 points per test image from SceneFlow in its clean renderings and compute the error of the prediction and the logarithm of the predicted scale of the Laplacian, \ie, the pixelwise uncertainty. On the left plot, we show the median of the error for a given predicted uncertainty (orange line), the 25- and 75-percentile in dark blue, and the 10- and 90-percentile in light blue. On the right plot, we sort pixels according to their predicted uncertainty from the less uncertain to the more uncertain and show the median of the error over the fractions of pixels considered (orange line),  the 25- and 75-percentile in dark blue, and the 10- and 90-percentile in light blue.}
\label{fig:error_vs_confidenc}
\end{figure*}

\vpara{Visualization of the uncertainty.} We visualize in Figure~\ref{fig:confidence} this uncertainty for a few examples. We observe that it is highly linked with the error of the predicted disparity as red areas in the error correspond to blue areas in the uncertainty maps.

\vpara{Statistics on the uncertainty.} To better measure the correlation of our predicted uncertainty with the error of the disparity prediction, we plot a few statistics in Figure~\ref{fig:error_vs_confidenc}. On the left one, we show some percentiles of the error when varying the predicted scale of the Laplacian distribution. We observe that a lower uncertainty clearly corresponds to pixels with lowest errors, while a high uncertainty corresponds to pixels with a higher error. On the right plot, we order pixels from the less uncertain to the more uncertain and show the percentiles of errors when increasing the ratio of pixels considered. We observe the same behavior, showing the correlation of our uncertainty with the error of the prediction. Note that 95\% of the pixels have an error below 1, thus the scale of the y-axis of the plot.

\begin{table}
\centering
\resizebox{\linewidth}{!}{
\begin{tabular}{lp{0cm}rrrrp{0cm}rrrr}
     \toprule
\multirow{2}{*}{loss} & & \multicolumn{4}{c}{Stereo (bad@1.0px$\downarrow$)} & & \multicolumn{4}{c}{Flow (EPE$\downarrow$)}  \\
 & & \multicolumn{1}{c}{\small Md} & \multicolumn{1}{c}{\small ETH} & \multicolumn{1}{c}{\small SF(c)} & \multicolumn{1}{c}{\small SF(f)} & & \multicolumn{1}{c}{\small FT(c)} & \multicolumn{1}{c}{\small FT(f)} & \multicolumn{1}{c}{\small Si.(c)} & \multicolumn{1}{c}{\small Si.(f)}  \\
\cmidrule(lr){1-1} \cmidrule(lr){3-6} \cmidrule(lr){8-11}
L1 & & 23.0 &  0.95 &   6.1 &   6.3 &    &   3.02 &  2.69 &  1.51 &  2.13   \\
\bf{Lap.} & &  \bf{15.5} &  \bf{0.38} &   \bf{5.0} &   \bf{5.3} &    &   \bf{2.85} &  \bf{2.45} &  \bf{1.43} &  \bf{1.99} \\
\bottomrule
\end{tabular}
} 
\vspace{-0.3cm}
\caption{\textbf{Impact of the loss.} We compare a standard L1 loss \vs the Laplacian (Lap.) loss.}
\label{tab:loss}
\end{table}

\vpara{Comparison with an L1 loss.} 
In Table~\ref{tab:loss}, we quantitatively evaluate the effect of using a loss on a Laplacian distribution (Equation~\ref{eq:loss}) compared to using only an L1 loss. In the latter case, we cannot leverage the predicted scale of a Laplacian distribution for merging overlapping tiles. We thus follow~\cite{flowformer} and use a weights that decrease with the distance to the center of the image. We observe that the Laplacian loss 
 outperforms the L1 loss on all stereo and flow benchmarks. A Laplacian loss can be interpreted as an L1 term, weighted for each pixel according to an uncertainty measure, thus allowing to downweight uncertain pixels in practice. In addition, having access to the scale of the Laplacian allows a more elegant merging strategy for the overlapping tiles.

\subsection{Towards smarter tiling}

One limitation of our approach mentioned in the main paper is the tiling-based inference.
For instance, \ours is based on crops with a width of 704 pixels, this means that for large disparity values, the matching pixels would be out of the scope of the corresponding tile in the second image. 
As an alternative, we have tried a strategy where a second tile in the second image is also considered, which is shifted by 150 pixels, thus reducing the disparity value by the same amount. With the model with ViT-Base encoder and Base decoder, such a strategy allows to reduce the bad@1.0 from 17.1\% to 12.0\% on Middlebury v3 validation set, when replacing the predictions over 200px from the original tile, with the ones from the secondary tile. While this strategy seems promising, it is however not really satisfactory as it multiples the number of tiles to proceed by 2. We hope to find better strategies in the future.

\begin{table*}
\centering
\resizebox{\linewidth}{!}{
\begin{tabular}{lr@{}rl}
\toprule
 stereo dataset & \multicolumn{2}{c}{\# pairs} & comment \\
 \midrule
 CREStereo~\cite{crestereo} &                            & 200,000 & all training pairs \\
 SceneFlow\cite{sceneflow} &                            &  70,908 & Driving, Monkaa and FlyingThings in both clean and final renderings \\
           &                  &         & ~~4,370 validation pairs from FlyingThings test for each rendering (clean and final) \\
 ETH3D Low Res~\cite{eth3d} & 30${\times}$ & 24      & `delivery\_area\_3s', `electro\_3l' `playground\_3l' (3 pairs) are kept apart for validation \\
 Middlebury v3~\cite{middlebury} & 50${\times}$ & 14      & `Vintge' (1 pair) is kept apart for validation, we use the `full' resolution \\ 
 Middlebury 21 & 50${\times}$ & 335     & `traproom1' and `traproom2' are kept apart for validation (20 pairs) \\ 
 Middlebury 14 & 50${\times}$ & 132     & `Umbrella-umperfect' and `Vintage-perfect' are kept apart for validation (6 pairs) \\         
 Middlebury 06 & 50${\times}$ & 171     & `Rocks1' and `Wood2' are kept apart for validation (18 pairs) \\
 Middlebury 05 & 50${\times}$ & 45      & `Reindeer' is kept apart for validation (9 pairs) \\
 Booster~\cite{booster}       &              & 213     & only the `balanced' subset, `Vodka' and `Washer' sequences (15 pairs)  kept apart for validation \\
 \midrule
 total         &              & 306,691& \\ 
\bottomrule
\end{tabular}
} \\[-0.3cm]
\caption{\textbf{Overview of our stereo training data.} We indicate here the train/val split used for the ablations, as well as the number of training pairs. For ETH3D and Middlebury, we also consider multiple times each pair in each epoch.}
\label{tab:stereodata}
\end{table*}

\begin{table*}
\centering
\resizebox{\linewidth}{!}{
\begin{tabular}{lrrl}
\toprule
 flow dataset & \# pairs & prob. & comment \\
 \midrule
FlyingChairs~\cite{flownet} & 22,232 & 12\% & - \\
FlyingThings~\cite{sceneflow} & 80,604 & 40\% & 40,302 pairs for both `clean' and `final' renderings \\
             &     &            & ~~we use the same 1,024 validation pairs from the test set as~\cite{unimatch} \\
MPI-Sintel~\cite{sintel}   & 943 & 10\% & sequences `temple\_2' and `temple\_3' (98 pairs) are kept apart for validation \\
TartanAir~\cite{tartan}    & 306,268 & 38\% & - \\
 \midrule
 total         &  410,047 & 100\% &  \\ 
\bottomrule
\end{tabular}
} \\[-0.3cm]
\caption{\textbf{Overview of our flow training data.} We indicate here the train/val split used for the ablations, as well as the number of remaining training pairs. During training, we set a number of images per epoch and randomly sample them among the available datasets with the percentages shown in the column `prob.'.}
\label{tab:flowdata}
\end{table*}

\section{Training details}
\label{app:details}

\vpara{\ours training.}
We train \ours for 32 epochs using batches of 6 pairs of $704{\times}352$ crops. 
We detail the training/validation pairs we use for our ablations in Table~\ref{tab:stereodata}.
We use the AdamW optimizer~\cite{adamw} with a weight decay of $0.05$, a cosine learning rate schedule with a single warm-up epoch and a learning rate of $3.10^{-5}$. During training, we apply standard data augmentations: color jittering (asymmetrically with probably 0.2), random vertical flipping with probably $0.1$, random scaling with probability $0.8$ in the range $[2^{-0.2},2^{0.4}]$ and stretching (resize different along the $x$ and $y$ axis) with probability $0.8$ in the range $[2^{-0.2},2^{0.2}]$, and slightly jitter the right image with probability $0.5$. 
When submitting to the official leaderboards, we include the pairs that were kept apart from the training sets for validation into the training epochs.

\vpara{\oursflow training.} We train \oursflow for $240$ epochs of $30,000$ pairs each, randomly sampled from all available data, using batches of $8$ pairs of crops of size $384{\times}320$. 
We detail the training/validation pairs we use for our ablations in Table~\ref{tab:flowdata}. 
To better balance the datasets, we set the probability of choosing a random pair from these datasets, see Table~\ref{tab:stereodata}. 
We
use the AdamW optimizer, a weight decay of $0.05$, a cosine learning rate schedule with linear warm-up over $1$ epoch, and a base learning rate of $2.10^{-5}$. During training, we apply standard augmentations~\cite{unimatch}: random color jittering (asymmetrically with probably 0.2), random scaling with probably $0.8$ with a scale sampled in $[2^{-0.2},2^{0.5}]$ and stretching with probability $0.8$ in the range $[2^{-0.2},2^{0.2}]$. 

\end{document}

%% file: plot/tiling/styles.tex
\pgfplotsset{compat=newest}
\usepgfplotslibrary{fillbetween}

\newcommand{\oxmedc}{CornflowerBlue}
\newcommand{\oxhardc}{TealBlue}
\newcommand{\oxf}{MidnightBlue}

\newcommand{\pamedc}{WildStrawberry}
\newcommand{\pahardc}{RedViolet}
\newcommand{\paris}{RedViolet}

\newcommand{\valc}{PineGreen}

\newcommand{\ourscolor}{Maroon}
\newcommand{\howcolor}{Black}

\newcommand{\oxfmark}{o}
\newcommand{\parmark}{triangle}
\newcommand{\valmark}{square}

\newcommand{\bone}{CornflowerBlue}
\newcommand{\btwo}{MidnightBlue}
\newcommand{\bthree}{TealBlue}
\newcommand{\bfour}{Blue}

\newcommand{\leg}[1]{\addlegendentry{#1}}
\newcommand{\legtext}[1]{\addlegendimage{empty legend} \addlegendentry{#1}}

\tikzset{every mark/.append style={solid}}
\pgfplotsset{
	grid=both, width=\linewidth, try min ticks=5,
    legend cell align=left, 
    legend style={fill opacity=0.8},
	ylabel near ticks,
    xlabel near ticks,
    every tick label/.append style={font=\footnotesize},
}

\pgfplotsset{
    % oxmed/.style={thick, color=\oxmedc, mark=o},
    % oxhard/.style={thick, color=\oxhardc, mark=o},
    % pamed/.style={thick, color=\pamedc, mark=star}, pahard/.style={thick, color=\pahardc, mark=star},
    oxmed/.style={thick, color=\oxf, mark=o},
    oxhard/.style={thick, dashed, color=\oxf, mark=o},
    pamed/.style={thick, color=\paris, mark=star}, 
    pahard/.style={thick, dashed, color=\paris, mark=star},
    val/.style={thick, color=\valc, mark=x},
    oursoxf/.style={thick, color=\ourscolor, mark=\oxfmark},
    howoxf/.style={thick, dashed, color=\howcolor, mark=\oxfmark},
    ourspar/.style={thick, color=\ourscolor, mark=\parmark},
    howpar/.style={thick, dashed, color=\howcolor, mark=\parmark},
    oursval/.style={thick, color=\ourscolor, mark=\valmark},
    howval/.style={thick, dashed, color=\howcolor, mark=\valmark},
    numean/.style={thick, color=\ourscolor, mark=none},
    numin/.style={thick, color=gray, mark=none},
    wid/.style={thick, color=\ourscolor, mark=o, mark size=0.5pt},
    woid/.style={thick, densely dashdotted, color=RedOrange, mark=o, mark size=0.5pt},
    d2k1/.style={thick, color=CornflowerBlue, mark=\oxfmark},
    d2k2/.style={thick, color=MidnightBlue, mark=\oxfmark},
    d2k4/.style={thick, color=TealBlue, mark=\oxfmark},
    d2k8/.style={thick, color=Blue, mark=\oxfmark},
    d4c256/.style={thick, color=\ourscolor, mark=\oxfmark},
    d8c256/.style={thick, color=Green, mark=\oxfmark},
    b1c/.style={thick, color=\bone, mark=o},
    b2c/.style={thick, color=\btwo, mark=o},
    b3c/.style={thick, color=\bthree, mark=o},
    b4c/.style={thick, color=\bfour, mark=o},
    mae/.style={thick, color=Gray, mark=star},
    y1c/.style={thick, color=Goldenrod, mark=o},
    r1c/.style={thick, color=Maroon, mark=o},
    r2c/.style={thick, color=WildStrawberry, mark=o},
    r3c/.style={thick, color=RedViolet, mark=o},
    y1c/.style={thick, color=Goldenrod, mark=o},
    y2c/.style={thick, color=Apricot, mark=o},
    y3c/.style={thick, color=BurntOrange, mark=o},
    y3cp/.style={thick, color=BurntOrange, mark=o, mark size=8pt},
    yvert/.style={dashed, thick, color=BurntOrange, mark=o},
    yvertp/.style={dashed, thick, color=BurntOrange, mark=o, mark size=8pt},
    tt3/.style={thick, color=Maroon, mark=o},
    tt4/.style={thick, color=BurntOrange, mark=o},
    tt1/.style={thick, color=Maroon, mark=star},
    tt2/.style={thick, color=BurntOrange, mark=star},
    ttl1/.style={thick, color=Maroon},
    ttl2/.style={thick, color=BurntOrange},
    ttl3/.style={thick, mark=star,mark size=3pt},
    ttl4/.style={thick, mark=o},
    crcr/.style={thick, color=LimeGreen, mark=o},
    % tlcman 
    crcrp/.style={thick, color=Goldenrod, mark=o,mark size=2.5pt},
    crcap/.style={thick, color=BurntOrange, mark=diamond,mark size=2.5pt},
    maep/.style={thick, color=Maroon, mark=star, mark size=2.5pt},
    crca/.style={thick, color=Gray, mark=diamond},
    crcb/.style={thick, color=Gray, mark=o, mark size=2.5pt},
    crcc/.style={thick, color=Gray, mark=+,mark size=2.5pt},
    crcd/.style={thick, color=Gray, mark=star,mark size=2.5pt},
    crce/.style={thick, color=LimeGreen, mark=+,mark size=2.5pt},
    gvert/.style={dashed, thick, color=Gray,mark size=2.5pt},
    fin/.style={thick, color=Maroon, mark=diamond,mark size=2.5pt},
}

%% file: plot/tiling/plots.tex
\newcommand{\plotMd}[0]{
\begin{tikzpicture}
\begin{axis}[%
  width=1.1\linewidth,
  height=4cm, 
   xlabel={\normalsize Overlap ratio},
   xlabel style={yshift=4pt},
  minor tick num=1,
  ticklabel style = {font=\normalsize},
  legend pos=north east,
    % legend pos=outer north east,
    % xmode=log,
    title={\normalsize Middlebury v3 (bad@1.0px$\downarrow$)},
            title style={yshift=-5pt},
  ]
\input{plot/tiling/data}
    \addplot[fin]      table[x=overlap,  y=bad_md] \map;
    %\addplot[crcap]      table[x=overlap,  y=bad_sfclean] \map;
    %\addplot[crcap]      table[x=cf_epoch,  y=cf_pampjpe] \map;
    % \addplot[maep]      table[x=ca_epoch,  y=ca_pampjpe]  \map; 
     %\addplot[crca]      table[x=inet_epoch,  y=inet_pampjpe]  \map; 
      %\addplot[crcb]      table[x=dino_epoch,  y=dino_pampjpe]  \map; 
      %\addplot[crca]      table[x=mae_epoch,  y=mae_pampjpe]  \map; 
     %\addplot[crcd]      table[x=croco_epoch,  y=croco_pampjpe]  \map; 
      %\addplot[crce]      table[x=nomc_epoch,  y=nomc_pampjpe]  \map; 
      %\addplot[gvert] coordinates {(0, 58.93) (50, 58.93)};
\end{axis}
\end{tikzpicture}}

\newcommand{\plotSf}[0]{
\begin{tikzpicture}
\begin{axis}[%
  width=1.1\linewidth,
  height=4cm, 
   xlabel={\normalsize Overlap ratio},
   xlabel style={yshift=4pt},
  minor tick num=1,
  ticklabel style = {font=\normalsize},
  legend pos=north east,
    % legend pos=outer north east,
    % xmode=log,
    title={\normalsize SceneFlow clean (bad@1.0$\downarrow$)},
            title style={yshift=-5pt},
  ]
\input{plot/tiling/data}
    \addplot[fin]      table[x=overlap,  y=bad_sfclean] \map;
\end{axis}
\end{tikzpicture}}

\newcommand{\plotSintelf}[0]{
\begin{tikzpicture}
\begin{axis}[%
  width=1.1\linewidth,
  height=4cm, 
   xlabel={\normalsize Overlap ratio},
   xlabel style={yshift=4pt},
  minor tick num=1,
  ticklabel style = {font=\normalsize},
  legend pos=north east,
    % legend pos=outer north east,
    % xmode=log,
    title={\normalsize MPI-Sintel final (EPE$\downarrow$)},
            title style={yshift=-5pt},
  ]
\input{plot/tiling/data}
    \addplot[fin]      table[x=overlap,  y=epe_sintelf] \map;
    %\addplot[crcap]      table[x=overlap,  y=bad_sfclean] \map;
    %\addplot[crcap]      table[x=cf_epoch,  y=cf_pampjpe] \map;
    % \addplot[maep]      table[x=ca_epoch,  y=ca_pampjpe]  \map; 
     %\addplot[crca]      table[x=inet_epoch,  y=inet_pampjpe]  \map; 
      %\addplot[crcb]      table[x=dino_epoch,  y=dino_pampjpe]  \map; 
      %\addplot[crca]      table[x=mae_epoch,  y=mae_pampjpe]  \map; 
     %\addplot[crcd]      table[x=croco_epoch,  y=croco_pampjpe]  \map; 
      %\addplot[crce]      table[x=nomc_epoch,  y=nomc_pampjpe]  \map; 
      %\addplot[gvert] coordinates {(0, 58.93) (50, 58.93)};
\end{axis}
\end{tikzpicture}}

\newcommand{\plotTiles}[0]{
\begin{tikzpicture}
\begin{axis}[%
  width=1.1\linewidth,
  height=4cm, 
   xlabel={\normalsize Overlap ratio},
   xlabel style={yshift=4pt},
  minor tick num=1,
  ticklabel style = {font=\normalsize},
  legend pos=north east,
    % legend pos=outer north east,
    % xmode=log,
    title={\normalsize Nb. tiles$\downarrow$},
            title style={yshift=-5pt},
  ]
\input{plot/tiling/data}
    \addplot[fin]      table[x=overlap,  y=nbtiles] \map;
\end{axis}
\end{tikzpicture}}

\newcommand{\plotTime}[0]{
\begin{tikzpicture}
\begin{axis}[%
  width=1.1\linewidth,
  height=4cm, 
   xlabel={\normalsize Overlap ratio},
   xlabel style={yshift=4pt},
  minor tick num=1,
  ticklabel style = {font=\normalsize},
  legend pos=north east,
    % legend pos=outer north east,
    % xmode=log,
    title={\normalsize Time $\downarrow$},
            title style={yshift=-3pt},
  ]
\input{plot/tiling/data}
    \addplot[fin]      table[x=overlap,  y=time] \map;
\end{axis}
\end{tikzpicture}}

%% file: plot/tiling/data.tex
% overlap = [0.0,0.1,0.2,0.3,0.4,0.5,0.6,0.7,0.8,0.9]

% bad@1.0_mdeval3 = [30.37,30.37,29.35,23.58,21.26,21.55,22.29,20.20,20.59,18.42]

% bad@1.0_sfclean = [5.89,5.89,5.89,5.89,5.89,5.79,5.79,5.78,5.77,5.78]

% nbtiles_for_fullhd =  [12, 12, 16, 16, 20, 30, 42, 56, 120, 418]
% % or 
% time_in_s_for_fullhd = [0.66, 0.66, 0.88, 0.88, 1.1, 1.65, 2.31, 3.08, 6.6, 22.99]

\pgfplotstableread{
overlap   	bad_md      bad_sfclean epe_sintelc  epe_sintelf nbtiles       time
0.0    		23.1  		5.89   		1.73         2.29              12    	0.66 
0.1    		23.1  		5.89   		1.73         2.29              12    	0.66 
0.2    		23.6  		5.89   		1.51         2.10              16        0.88 
0.3    		22.6  		5.89   		1.51         2.10              16    	0.88 
0.4    		19.1  		5.89   		1.51         2.10              20    	1.1  
0.5    		17.7  		5.79   		1.52         2.11              30    	1.65 
0.6    		17.2  		5.79   		1.52         2.06              42    	2.31 
0.7    		16.2  		5.78   		1.44         2.01              56    	3.08 
0.8    		16.2  		5.77   		1.44         2.00             120   	6.6  
0.9    		15.5  		5.78   		1.43         1.9              418   	22.99
}{\map}               